\documentclass[10pt,twocolumn,letterpaper]{article}

\usepackage[pagenumbers]{wacv} 

\usepackage{times}
\usepackage{epsfig}
\usepackage{graphicx}
\usepackage{amsmath}
\usepackage{amssymb}
\usepackage{booktabs}

\usepackage{pifont}

\usepackage{color}
\usepackage{soul}
\usepackage{xcolor}
\usepackage{multirow}
\usepackage{booktabs}
\usepackage{enumerate}
\usepackage{enumitem}
\usepackage{pifont}%
\usepackage{caption}
\definecolor{RowColor}{rgb}{0.91, 0.91, 1}
\definecolor{babyblue}{rgb}{0.63, 0.79, 0.95}
\usepackage{colortbl}
\usepackage{wrapfig}

\newcommand{\cmark}{\ding{51}}%
\newcommand{\xmark}{\ding{55}}%
\usepackage[pagebackref,breaklinks,colorlinks]{hyperref}

\usepackage[capitalize]{cleveref}
\crefname{section}{Sec.}{Secs.}
\Crefname{section}{Section}{Sections}
\Crefname{table}{Table}{Tables}
\crefname{table}{Tab.}{Tabs.}

\def\wacvPaperID{456} 
\def\confName{WACV}
\def\confYear{2025}

\begin{document}

\title{DiffMesh: A Motion-aware Diffusion Framework for Human Mesh Recovery from Videos}

\author{Ce Zheng$^{1}$, Xianpeng Liu$^{2}$, Qucheng Peng$^{3}$, Tianfu Wu$^{2}$, Pu Wang$^{4}$,  Chen Chen$^{3}$\\
$^1$ Carnegie Mellon University, 
$^2$ North Carolina State University\\
$^3$ Center for Research in Computer Vision, University of Central Florida\\
$^4$ University of North Carolina at Charlotte
}

\twocolumn[{%
\renewcommand\twocolumn[1][]{#1}%
\maketitle
\begin{center}
\vspace{-20pt}
    \centering
   \includegraphics[width=0.91\textwidth]{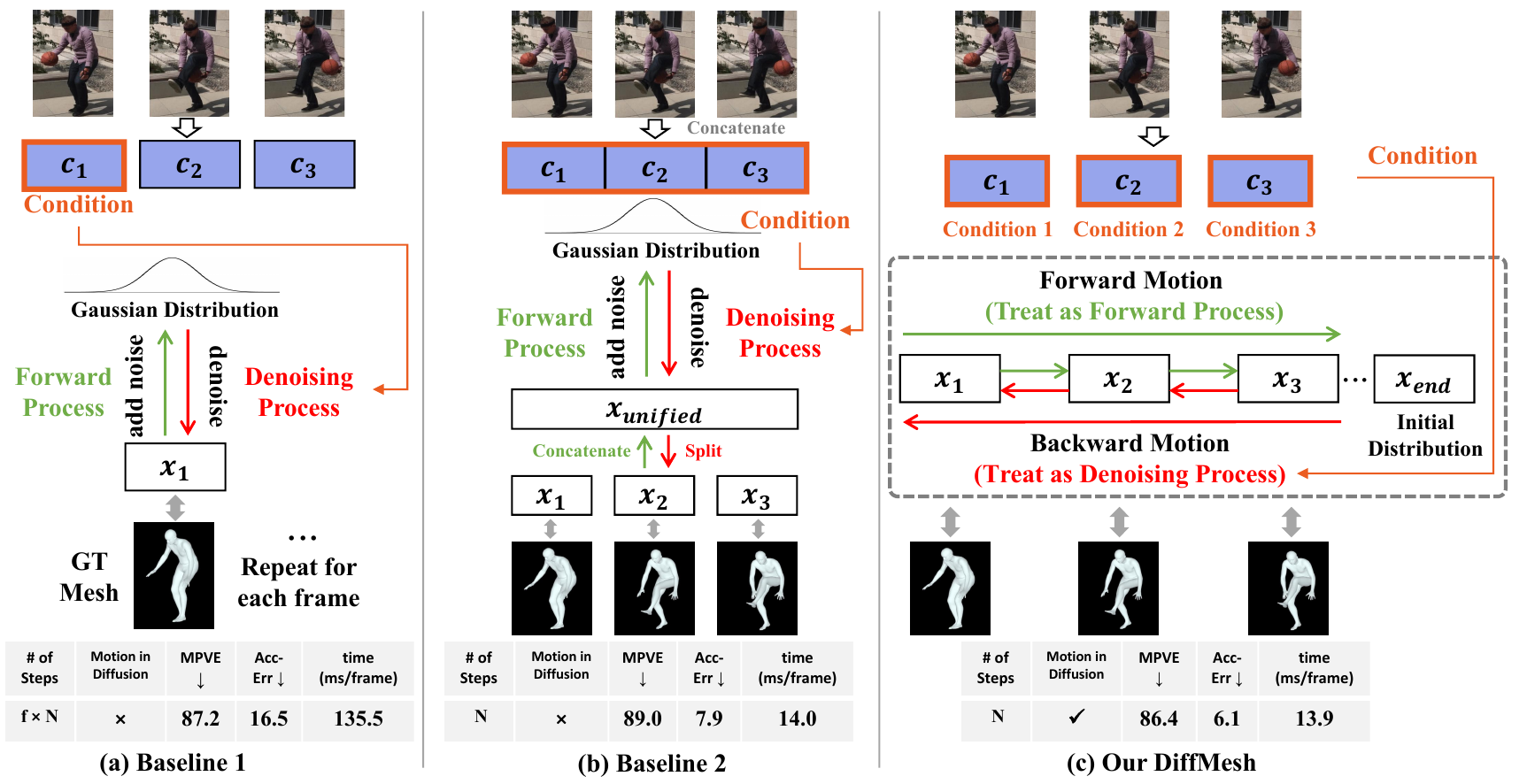}
   \vspace{-5pt}
    \captionof{figure}{Different approaches of applying diffusion model for video-based HMR, the input frame $f$ is 3 for simplicity and the number of steps is $N$. Here $x_{i}$ and $c_{i}$ denote the mesh and conditional features of $i_{th}$ frame, respectively. (a) Baseline 1: The diffusion model is applied for each frame individually. The total steps are $f \times N$, which is computationally expensive. Moreover, motion is ignored during the denoising process, leading to non-smooth motion predictions. (b) Baseline 2: The features from ground truth mesh feature of each frame $x_i$ are concatenated to unified features $x_{unified}$ during the forward process. To obtain the mesh of each frame, features are split after the denoising process. Although this strategy reduces the total steps to $N$, it doesn't effectively model human motion during denoising. (c) Our proposed DiffMesh: We consider the inherent motion patterns within the forward process and the reverse process when utilizing the diffusion model, leading to smooth and accurate motion predictions. The total number of steps remains as $N$. For more detailed information, please refer to Sec \ref{DiffMesh}.}
    \label{fig:pipe_comp}
\end{center}%
}]

\begin{abstract}
   \vspace{-5pt}
   Human mesh recovery (HMR) provides rich human body information for various real-world applications such as gaming, human-computer interaction, and virtual reality. While image-based HMR methods have achieved impressive results, they often struggle to recover humans in dynamic scenarios, leading to temporal inconsistencies and non-smooth 3D motion predictions due to the absence of human motion. In contrast, video-based approaches leverage temporal information to mitigate this issue. In this paper, we present DiffMesh, an innovative motion-aware diffusion framework for video-based HMR. DiffMesh establishes a bridge between diffusion models and human motion, efficiently generating accurate and smooth output mesh sequences by incorporating human motion within the forward process and reverse process in the diffusion model. Extensive experiments are conducted on the widely used datasets (Human3.6M \cite{h36m_pami} and 3DPW \cite{pw3d2018}), which demonstrate the effectiveness and efficiency of our DiffMesh. Visual comparisons in real-world scenarios further highlight DiffMesh's suitability for practical applications. \textcolor{magenta}{The project webpage is: \url{https://zczcwh.github.io/diffmesh_page/} }
\end{abstract}

\vspace{-10pt}
\section{Introduction}
\label{sec:intro}
Human Mesh Recovery (HMR) aims to recover detailed human body information, encompassing both pose and shape, with applications spanning gaming, human-computer interaction, and virtual reality  \cite{hmrsurvey}. While image-based HMR methods have achieved remarkable results, they often fall short in dynamic scenarios, wherein the neglect of human motion results in temporal inconsistencies and non-smooth 3D motion predictions. In contrast, video-based approaches offer a solution to mitigate these issues by leveraging temporal cues from input sequences, thereby meeting the practical demands of real-world applications. This makes video-based methods a compelling choice for enhancing the temporal coherence and consistency of 3D motion predictions.

Recently, diffusion models have garnered substantial interest within the computer vision community.  By learning to reverse the diffusion process in which noise has been added in successive steps, diffusion models can generate samples that match a specified data distribution corresponding to the provided dataset. As generative-based methods, diffusion models excel in tasks such as motion synthesis and generation \cite{MDM,flame,li2023ego}. The quality and diversity of generated outputs have witnessed significant improvements by bridging the gap between uncertain and determinate distributions. This inspires us to consider the diffusion model as a promising solution for recovering high-quality human mesh from input data fraught with depth ambiguity.

While it is straightforward to apply the diffusion model for image-based HMR such as in \cite{foo2023distribution}, effective utilization of the diffusion model for video-based HMR remains a formidable challenge. As previous video-based HMR methods have not yet incorporated diffusion models into their frameworks, we provide a summary of diffusion-based methods related to human pose and mesh recovery (please refer to Sec. \ref{sec:related_diff} for further elaboration). We then organize these methods into two intuitive baseline approaches that can be applied for video-based HMR, as illustrated in Fig. \ref{fig:pipe_comp} (a) and (b).
Assuming a video input with $f$ frames in \ul{Baseline 1}, the diffusion model is applied individually to recover the mesh of the single frame (usually would be the center frame). More details can be found in Sec. \ref{Baseline}.
However, this approach suffers from significant computational burdens as the total number of denoising steps amounts to  $f \times N$ to recover the mesh sequence of $f$ frames, with $N$ representing the number of denoising steps in the diffusion model. Moreover, the diffusion model does not account for motion smoothness in dynamic sequences, which may result in potentially temporal inconsistent mesh predictions. To mitigate computational costs and enhance inference speed, we introduce \ul{Baseline 2}. The mesh feature of each frame $x_{i}$, where $i \in \left\{ 1,\dots f \right\}$ are concatenated to unified features $x_{unified}$ before the forward diffusion process. 
Subsequently, the output features are partitioned back into individual frames $x_{i}$ after the \textit{denoising process}. 
While this strategy effectively reduces the steps from $f \times N$ to $N$,  it is important to note that motion smoothness in dynamics is still not considered during the forward and reverse process since features across frames are simply concatenated. This limitation can potentially lead to non-smooth motion predictions.

\begin{figure}[htp]
  \centering
  \vspace{-10pt}
  \includegraphics[width=1\linewidth]{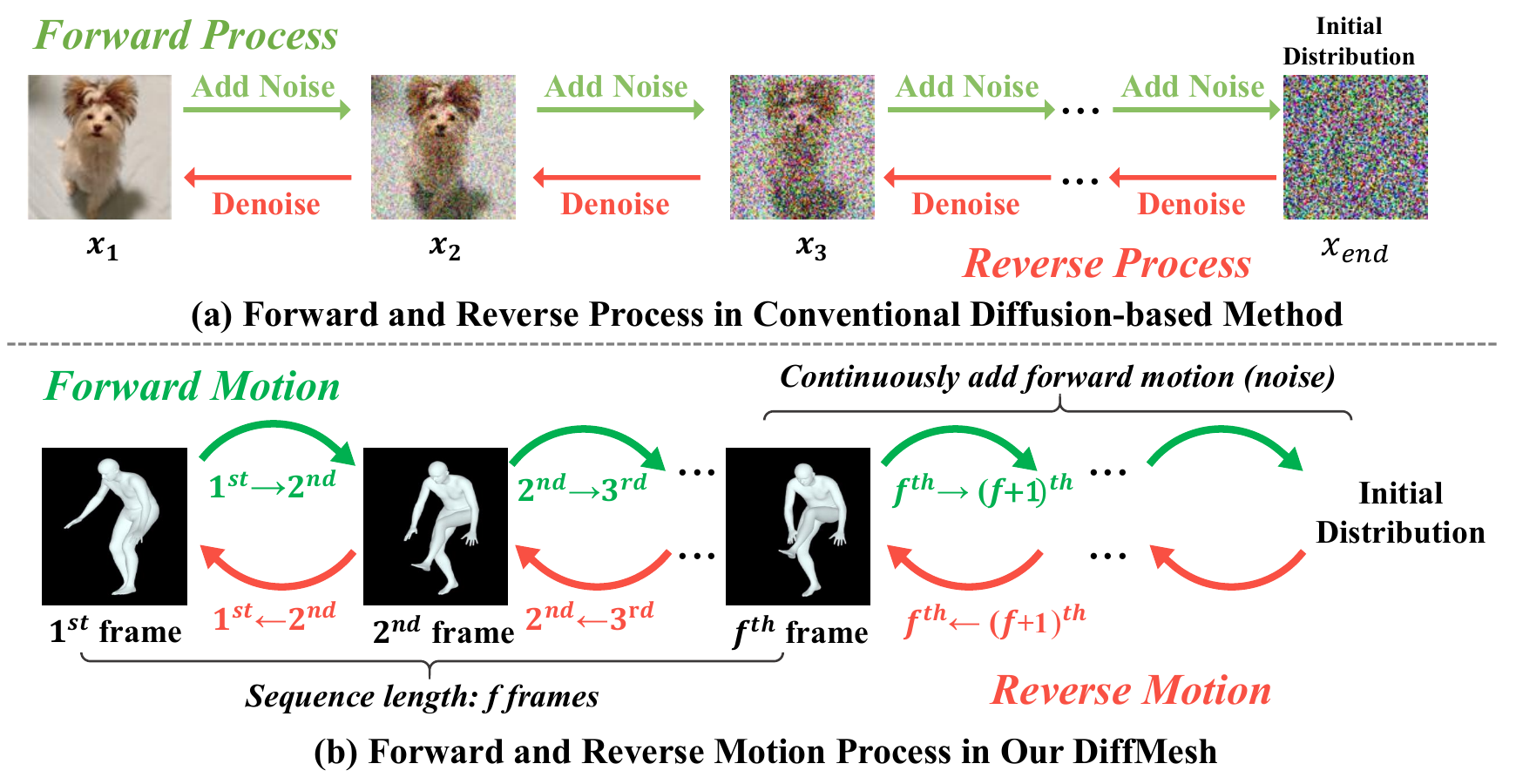}
  \vspace{-15pt}
  \caption{(a) The general pipeline for diffusion model. Input data is perturbed by adding noise recursively and output data is generated from the noise in the reverse process. Images are taken from \cite{diffsurvey1}. (b) Human motion is involved over time in the input video sequence. Similar to the forward process in (a), the forward motion between adjacent frames resembles the process of adding noise. The mesh of the previous frame can be decoded through the reverse motion process successively. }
  \label{fig:dm_tid}
  \vspace{-15pt}
\end{figure}

To tackle the aforementioned efficiency and motion smoothness challenges when utilizing the diffusion model, we introduce \textbf{DiffMesh}, a novel motion-aware diffusion framework as illustrated in Fig. \ref{fig:pipe_comp} (c). DiffMesh is designed specifically for video-based HMR. Unlike conventional diffusion models, which necessitate the addition of Gaussian noise from the input image during the \textit{forward diffusion process}, as depicted in Fig. \ref{fig:dm_tid} (a), DiffMesh adopts a distinct approach. We consider the forward motion influence over a brief time interval as shown in Fig. \ref{fig:dm_tid} (b), encompassing the features (embedded in the high-dimensional space) of adjacent frames $x_i$ to $x_{i+1}$, similar to the mechanism of introducing noise. Importantly, our approach does not conclude the forward motion at the last frame $x_{f}$, where $f$ is the total number of frames. 
Instead, we assume that 
the initial distribution can be reached after continuously adding forward motion by finite diffusion steps, which is similar to the  Gaussian distribution in the conventional diffusion model. 
During the \textit{reverse process}, the diffusion model decodes the previous feature $x_{i-1}$ based on $x_{i}$ and the corresponding conditional feature $c_{i}$ from the input frame. Consequently, the total number of steps remains consistent at $N$, as illustrated in Fig. \ref{fig:pipe_comp} (b). 

\textbf{Contribution:} The key innovation of DiffMesh lies in its capacity to consider the inherent motion patterns within the diffusion model efficiently and effectively. This consideration notably enhances motion smoothness and temporal consistency throughout the output mesh sequence, achieved through successive steps across frames. Incorporating our diffusion framework, we design a two-stream transformer architecture to decode motion features. Our DiffMesh not only achieves new state-of-the-art results on the Human3.6M \cite{h36m_pami} and 3DPW \cite{pw3d2018} datasets, but also demonstrates its superiority in handling in-the-wild scenarios compared to previous methods. These advantages make DiffMesh more viable for deployment in real-world applications for generating accurate and smooth human mesh sequences efficiently.

\section{Related Work}
Since human mesh recovery is a popular topic in computer vision, here we focus on the most relevant works to our methods. More references can be found in the \textcolor{blue}{supplementary Sec. 2}. We also refer readers to the recent and comprehensive HMR survey \cite{hmrsurvey} and diffusion model survey \cite{diffsurvey1} for more details.

\subsection{Video-based HMR}

Image-based HMR methods ~\cite{lin2021metro,zheng2023feater,li2023hybrik,zheng2023potter,zhang2023pymafx,li2022cliff,goel2023humans,wang2023refit,kaufmann2023emdb} have achieved remarkable progress in accuracy.  
However, when applied to video sequences, these image-based methods suffer from severe motion jitter due to frame-by-frame reconstruction, making them unsuitable for practical use. To address this issue, video-based methods \cite{hmmr,meva,sun2019human,zeng2022deciwatch,li2022dnd,GLoT,you2023coeval,yang2023capturing,MotionBERT2022,tian2024vita} utilize temporal information to enhance accurate and temporally consistent human mesh from video frames.
VIBE \cite{kocabas2020vibe} leverages the AMASS dataset \cite{AMASS2019} to discriminate between real human motions and those produced by its temporal encoder. MEVA \cite{meva} estimates the coarse 3D human motion using a variational motion estimator, then refines the motion by a motion residual regressor. TCMR \cite{tcmr} utilizes GRU-based temporal encoders to forecast the current frame from the past and future frames, which enhances the temporal consistency. Similar to \cite{tcmr}, MPS-Net \cite{MPS-Net} captures the motion continuity dependencies by the attention mechanism. MotionBERT \cite{MotionBERT2022} proposes a pertaining stage to recover 3D motion from noise 2D observations, achieving state-of-the-art performance.

\subsection{Diffusion Generative Models}
\label{sec:related_diff}
Diffusion Generative Models have achieved impressive success in a wide variety of computer vision tasks such as image inpainting \cite{song2021scorebased}, text-to-image generation \cite{photorealistic}, and image-to-image translation \cite{choi2021ilvr}. Given the strong capability to bridge the large gap between highly uncertain and determinate distribution, several works have utilized the diffusion generative model for the text-to-motion generation \cite{modiff,MDM,zhang2022motiondiffuse,li2025controlnet}, human pose estimation \cite{shan2023diffusion}, object detection \cite{chen2023diffusiondet}, and head avatar generation \cite{bergman2023articulated,mendiratta2023avatarstudio}. 
DiffPose \cite{gong2023diffpose} is the first to utilize a diffusion-based approach for predicting the 3D pose from a 2D pose sequence input. Shan et al. \cite{shan2023diffusion} generate multiple hypotheses of the 3D pose from 2D pose input using a diffusion model, and then a multi-hypothesis aggregation module outputs the final 3D pose. EgoEgo \cite{li2023ego} employs a diffusion model to generate multiple plausible full-body motions based on the head pose input. 

\begin{table}[htp]
\renewcommand\arraystretch{1.1}
\vspace{-5pt}
  \centering
  \resizebox{1\linewidth}{!}
  {
\begin{tabular}{l|c|ccc|c|c}
\hline
\multicolumn{1}{c|}{Methods} & Input              & \multicolumn{3}{c|}{Output}    & Motion-Aware   & Approach            \\ \cline{3-5}
\multicolumn{1}{c|}{}        &                    & 3D Pose & 3D Mesh & multi-frame  & Diffusion &    Similar to                   \\ \hline
DiffPose \cite{gong2023diffpose}       & 2D Human Pose Sequence   & \cellcolor{babyblue} \cmark     & \xmark      & \xmark   & \xmark        & Baseline 1 \\
Diff3DHPE \cite{zhou2023diff3dhpe}      & 2D Human Pose Sequence   & \cellcolor{babyblue} \cmark    & \xmark    & \cellcolor{babyblue} \cmark        & \xmark    & Baseline 2 \\
D3DP \cite{shan2023diffusion}           & 2D Human Pose Sequence   & \cellcolor{babyblue} \cmark     & \xmark      & \cellcolor{babyblue} \cmark  & \xmark        & Baseline 2 \\
HMDiff \cite{foo2023distribution}          & Single Image       & \cellcolor{babyblue} \cmark     & \cellcolor{babyblue} \cmark     & \xmark       & \xmark    & Baseline 1 \\
EgoEgo  \cite{li2023ego}                     & Head Pose Sequence & \cellcolor{babyblue} \cmark     & \cellcolor{babyblue} \cmark     & \cellcolor{babyblue} \cmark    & \xmark      & Baseline 2 \\ \hline
DiffMesh (ours)              & Video sequence     & \cellcolor{babyblue} \cmark     & \cellcolor{babyblue} \cmark     & \cellcolor{babyblue} \cmark        &  \cellcolor{babyblue} \cmark  &   \textcolor{red}{Our design}                    \\ \hline
\end{tabular}
}
\vspace{-5pt}
\caption{Comparison with previous diffusion-based human pose and mesh methods. }
\label{tab:frame-comp}
\vspace{-10pt}
\end{table}

\noindent \textbf{Distinction of Our Method:}
Previous methods for video-based HMR have not integrated diffusion models into their frameworks. While there are diffusion-based methods that can output 3D human pose or human mesh, they are not directly suitable for the video-based HMR task and require adaptations. We organize them into two intuitive baseline approaches. As depicted in Table \ref{tab:frame-comp}, approaches similar to \ul{Baseline 1}, such as \cite{SpatioTemporalDiffPose,foo2023distribution,gong2023diffpose}, require $f \times N$ denoising steps to produce output for $f$ frames. Conversely, methods similar to \ul{Baseline 2}, such as \cite{zhou2023diff3dhpe,shan2023diffusion,li2023ego}, reduce the total denoising steps from $f \times N$ to $N$. 
However, these methods do not effectively incorporate the consideration of human motion within the sequence during both the forward process and the reverse process, which may potentially result in temporal inconsistencies and non-smooth 3D motion predictions. 
In contrast, our approach treats motion patterns across frames as the inherent noise in the forward process.
This unique motion-aware design in diffusion enables our approach to decode human motion patterns during reverse processes within $N$ steps, setting it apart from conventional diffusion-based methods.

\section{Methodology}
\subsection{Preliminary}
\label{Preliminary}
We first provide a brief overview of the original denoising diffusion probabilistic model \cite{ho2020denoising} (DDPM) scheme. As shown in Fig. \ref{fig:dm_tid} (a), the DDPM consists of two Markov chains, which are a diffusion process and a reverse process. The forward chain perturbs data to noise, while the reverse chain converts the noise back into data. For a more detailed understanding, we direct interested readers to the original paper  \cite{ho2020denoising} and recent surveys of the diffusion model \cite{diffsurvey1,diffsurvey2}.

\noindent \textbf{Forward Diffusion Process:} Let  $x_0  \sim p(x_0)$ where  $x_0$ is the training sample and $ p(x_0)$ be the data density. Gaussian transitions $\mathcal{N}$ are successively added to perturb the training sample $x_0$. A sequence of corrupted data $x_1, x_2, \dots, x_T$ can be obtained following the forward Markov chain:
\vspace{-5pt}
\begin{align}
\small
\resizebox{0.9\linewidth}{!}{$
    p(x_t | x_{t-1}) = \mathcal{N}(x_t; \sqrt{1-\beta_t} \cdot x_{t-1}, \beta_t \cdot \textbf{I}), \forall t \in \left\{ 1,\dots T \right\}
\vspace{-10pt}
$}
\end{align}
where $\beta_1 \dots \beta_T \in \left [0,1\right )$ are the hyperparameters for the variance schedule in each step, $T$ is the number of steps, and $\textbf{I}$ is the identity matrix. We use $\mathcal{N}(x;\mu,\sigma^2)$ to denote the normal distribution of mean $\mu$ and standard deviation $\sigma$ that produce $x$.  According to the property of normal distribution recursive formulation, we can directly sample $x_t$ when $t$ is drawn from a uniform distribution as follows: 
\vspace{-5pt}
\begin{align}
\small
    p(x_t | x_{0}) = \mathcal{N}(x_t; \sqrt{\hat{\beta_t}} \cdot x_{0}, (1-\hat{\beta_t}) \cdot \textbf{I}), 
\vspace{-10pt}
\end{align}
where $\hat{\beta_t} = \prod_{i=1}^{t} \alpha_i $ and $\alpha_t = 1- \beta_t$.

Based on this equation, we can sample any $x_t$ directly when given original data $x_0$ and a fixed variance schedule $\beta_t$. To achieve this, the sample $x$ of a normal distribution $x \sim \mathcal{N}(\mu , \sigma^2 \cdot \textbf{I} )$  first subtracts the mean $\mu$ and divides by the standard deviation $\sigma$, outputting in a sample $z= \frac{x-\mu}{\sigma}$ of the standard normal distribution $z \sim \mathcal{N}(0, \textbf{I})$. Thus  $x_t$ is sampled from $p(x_t | x_0)$ as follows: 
\vspace{-5pt}
\begin{align}
\small
    x_t = \sqrt{\hat{\beta_t}} \cdot x_{0} + \sqrt{(1-\hat{\beta_t})} \cdot z_t, 
\vspace{-10pt}
\end{align}
where $z_t \sim \mathcal{N}(0,\textbf{I})$. 

\noindent \textbf{Reverse Denoising Process:} To generate new samples from $p(x_0)$, we apply the reverse Markov chain from a $x_T \sim \mathcal{N}(0, \textbf{I})$: 
\vspace{-5pt}
\begin{align}
\small
    p_{\theta}(x_{t-1} | x_t) = \mathcal{N}(x_{t-1}; \mu_{\theta}(x_t, t), \Sigma_{\theta}^{}(x_t, t) ), 
\label{eq:reverse}
\end{align}
where $\theta$ denotes model parameters, and the mean $\mu_{\theta}(x_t, t)$ and covariance $\Sigma_{\theta}(x_t, t) $ are parameterized by deep neural networks $\epsilon_\theta$. 
\vspace{-5pt}
\begin{align}
\small
    x_{t-1} = \frac{1}{\sqrt{\beta_{t}}} (x_t - \frac{1-\beta_t}{\sqrt{1-\hat{\beta_t}}} \epsilon_\theta(x_t, t))+\sigma_t z, 
\label{eq:reverse_xt}
\vspace{-5pt}
\end{align}
If $t>1, z \sim \mathcal{N}(0,\textbf{I})$, else $z = 0$.

\subsection{Diffusion-based Baselines}
\label{Baseline}


\begin{wrapfigure}{r}{0.25\textwidth}
\vspace{-10pt}
  \centering
  \includegraphics[width=1\linewidth]{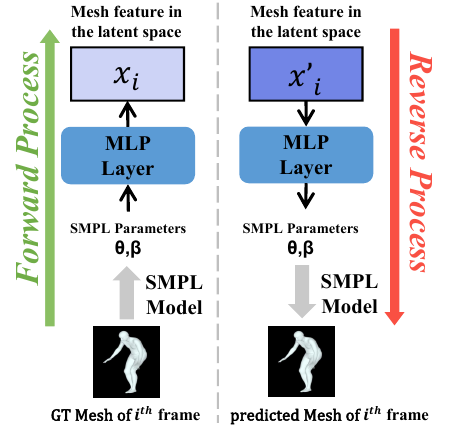}
  \vspace{-20pt}
  \caption{Transformation between human mesh and corresponding mesh feature in the latent space.}
  \label{fig:encoder}
  \vspace{-5pt}
\end{wrapfigure}
Given that previous video-based HMR methods have not integrated diffusion models into their frameworks, we incline to explore the performance of simply incorporating diffusion models in the context of video-based HMR. The forward and reverse processes are conducted in a latent space, the transformation between human mesh and the corresponding mesh feature in this latent space is shown in Fig. \ref{fig:encoder}.

\begin{figure}[htp]
\vspace{-5pt}
  \centering
  \includegraphics[width=0.95\linewidth]{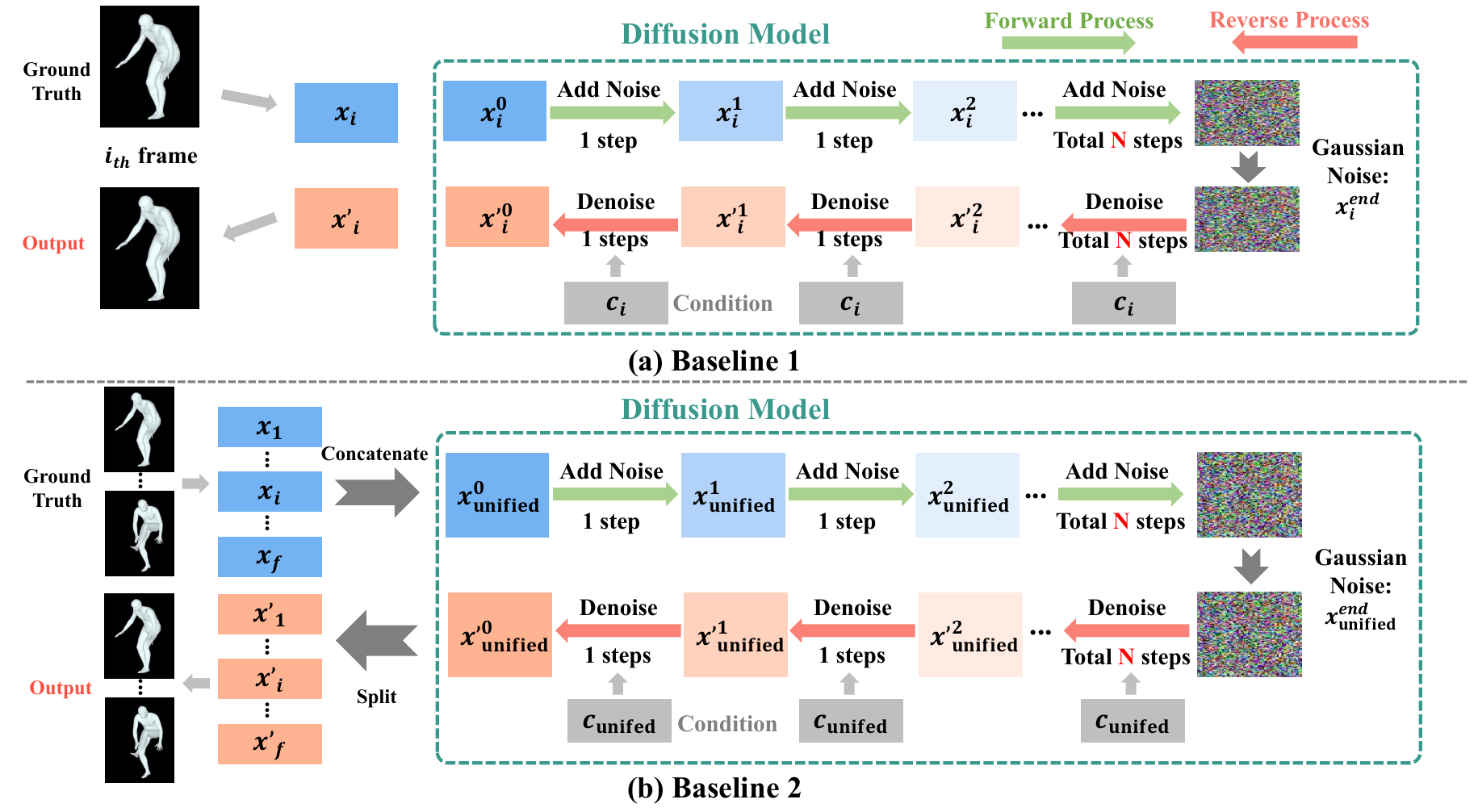}
  \vspace{-10pt}
  \caption{Two diffusion baselines for video-based HMR.}
  \label{fig:baseline}
  \vspace{-20pt}
\end{figure}

\noindent \textbf{Baseline 1:} As illustrated in Fig. \ref{fig:baseline} (a), we initially apply the vanilla diffusion model at the frame level. For each $i_{th}$ frame, we employ a backbone to extract the conditional feature $c_{i}$ where $i\in \left\{ 1,\dots, f \right\}$ and $f$ representing the total number of frames. We then proceed with the standard forward diffusion process, introducing noise to the mesh feature $x_{i}$ from ground truth SMPL parameters. After $N$ steps, the feature $x_{i}^{end}$ converges to a  Gaussian Noise. Subsequently, a transformer-based diffusion model  (details are included in the \textcolor{blue}{supplementary Sec. 2}) is employed for the denoising process. The denoised feature  $x'_{i}$ is obtained after $N$ denoising steps. Finally, the human mesh of $i_{th}$ frame is returned by the human mesh head. By iterating this procedure $f$ times, we generate the final human mesh sequence consisting of $f$ frames. However, despite its simplicity and straightforwardness, Baseline 1 exhibits a significant drawback: the total number of steps is  $f \times N$, resulting in considerable computational overhead and slow inference speed. Furthermore, it fails to address motion smoothness within the diffusion model, potentially leading to temporal inconsistencies in mesh predictions.

\noindent \textbf{Baseline 2:} To mitigate computational costs and enhance inference speed, we introduce Baseline 2 for video-based HMR.  Motivated by recent works \cite{modiff,flame,li2023ego} that leverage diffusion models for motion generation, we adopt a similar concatenation and separation strategy, as illustrated in Fig. \ref{fig:baseline} (b). In this approach, we concatenate the ground-truth mesh feature of each frame into a unified feature $x_{unified} = cat(x_1, x_2, \dots, x_f)$, where $f$ represents the total number of frames. Similarly, we create a concatenated conditional features  $c_{unified}$ by combining the image features from each frame. Subsequently, we apply the forward diffusion process by introducing noise for $N$ steps, which is analogous to Baseline 1. After $N$ steps,  $x_{unified}^{end}$ also converges to a Gaussian Noise state. Following this, we perform $N$ denoising steps utilizing a transformer-based diffusion model (details are included in the \textcolor{blue}{supplementary  Sec. 2}). The denoised feature $x'_{unified}$ is then partitioned into  $x'_1, x'_2, \dots, x'_f$. Finally, the mesh sequence is obtained through the mesh head.

\begin{figure*}[htp]
\vspace{-10pt}
  \centering
  \includegraphics[width=1\linewidth]{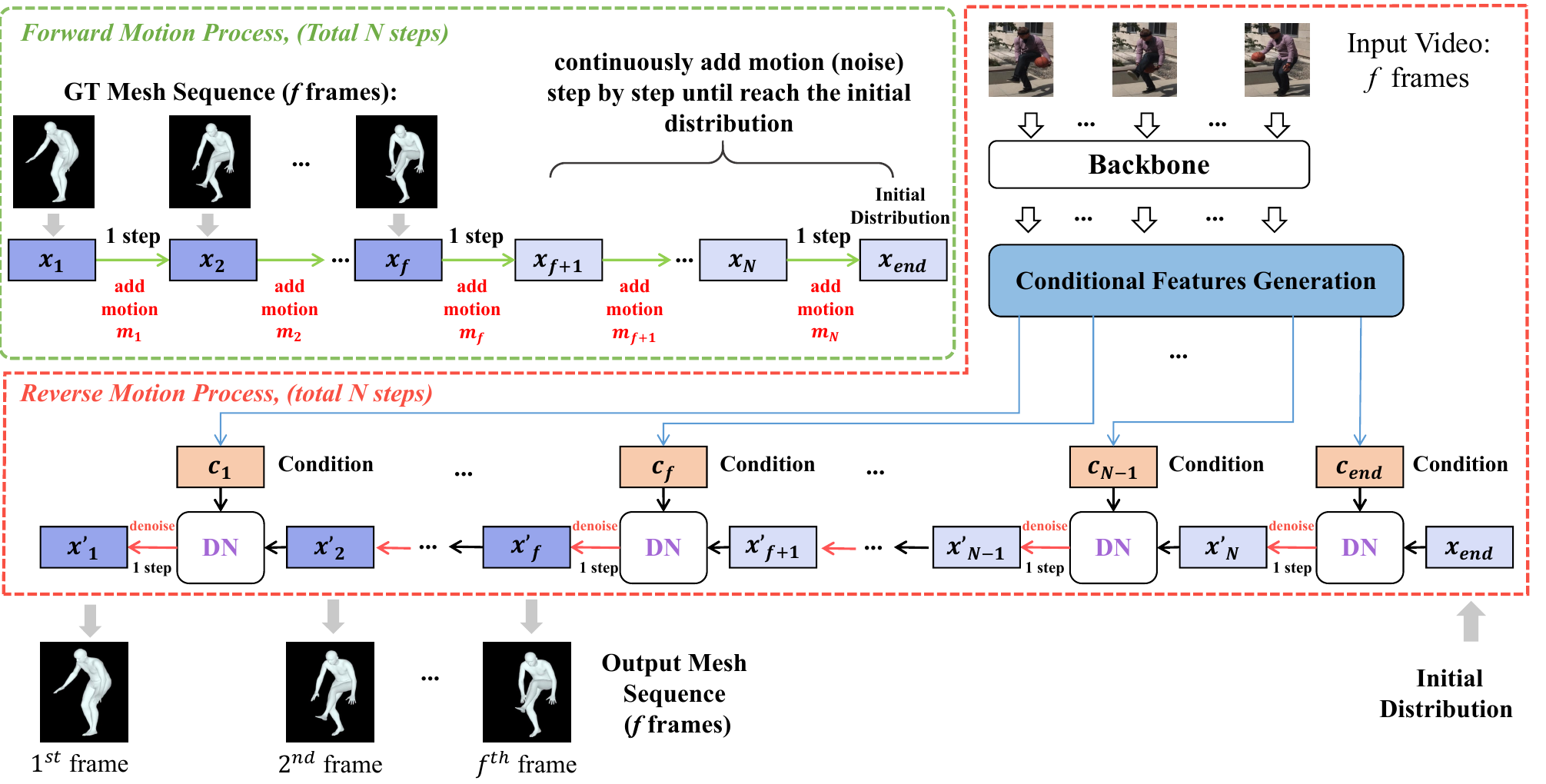}
  \vspace{-10pt}
  \caption{The architecture of DiffMesh: Our framework takes input sequence with $f$ frames, with the objective of outputting a human mesh sequence consisting of $f$ frames. We model the forward human motion across frames similar to the mechanism of introducing noise in the forward process. We assume that the human motion sequence will eventually reach the initial distribution within total $N$ steps. 
  (additional $N - f + 1$ steps are necessary from $x_f$ to reach the initial distribution $x_{end}$). Consequently, we utilize a transformer-based diffusion model to sequentially produce the decoded features during the reverse process. The final human mesh sequence is returned by a mesh head using SMPL \cite{SMPL:2015} human body model. The structure of DN (diffusion network) is illustrated in Fig. \ref{fig:md}. }
  \label{fig:arch}
  \vspace{-15pt}
\end{figure*}

The concatenation and separation strategy has substantially reduced the number of steps from $f \times N$ to $N$. However, it is important to note that this strategy does not effectively consider motion patterns across frames due to simple concatenation, which may result in temporal inconsistencies and non-smooth 3D mesh predictions.


\subsection{DiffMesh}
\label{DiffMesh}

To address the previously mentioned limitations in our two baselines, we present DiffMesh, a novel solution for video-based HMR. The overall architecture is depicted in Fig. \ref{fig:arch}. Our framework takes input sequence with $f$ frames, with the objective of outputting a human mesh sequence consisting of $f$ frames. Additionally, we extract the conditional feature $c_i$ using the backbone, incorporating a conditional features generation block (more details can be found in the \textcolor{blue}{supplementary  Sec. 2}). In contrast to Baseline 1 and Baseline 2, DiffMesh employs a distinct approach that accounts for the forward and backward motion within the diffusion model.

\noindent \textbf{Forward Process:}
We make the assumption that human motion consistently influences the human mesh feature (in a latent space) across adjacent frames, denoted as $x_{i}$ to $x_{i+1}$,  where $ i \in \left\{ 1,\dots f-1 \right\}$. This motion pattern can be conceptualized as a specific noise introduced to the mesh feature of the preceding frame: 
\vspace{-5pt}
\begin{align}
\small
    x_{i+1} = \sqrt{\beta_{i}} \cdot x_{i} + \sqrt{(1-\beta_{i})} \cdot \textcolor{orange}{m_{i}},
\label{eq:diffmesh_xi}
\vspace{-5pt}
\end{align}
where \textcolor{orange}{$m_{i}$} represents the motion pattern from $i_{th}$ frame to $(i+1)_{th}$ frame. Consequently,  the mesh feature of the first frame $x_{1}$  can be gradually diffused into the mesh feature of the last frame $x_{f}$ after $(f-1)$ forward diffusion steps. Importantly, achieving the last frame $x_{f}$ does not denote the conclusion of the forward diffusion process. \textit{We assume that mesh feature with adding human motion will eventually reach the initial distribution $x_{end}$ through total $N$ diffusion steps (similar to the total $N$ steps to get the Gaussian distribution in the conventional diffusion model)}.  
\ul{More details such as mathematical proof and ablation study are provided in the} \textcolor{blue}{supplementary  Sec. 3}.
To arrive at this initial distribution $x_{end}$, DiffMesh requires  $N-f+1$ additional forward diffusion steps from $x_f$. Therefore, the total number of steps remains $N$.

\noindent \textbf{Reverse Process:} Our objective during the reverse process is to obtain output features  $x'_{1} \cdots x'_{f}$. Initiating from the initial distribution  $x_{end}$, we design a transformer-based diffusion model, as depicted in Fig. \ref{fig:arch}, to sequentially produce decoded features. After  $N-f+1$ steps, our diffusion model yields $x'_{f}$, which serves as the target feature for the final frame $f$. Iteratively, we generate the output features $x'_{i}$ where $ i \in \left\{ f-1,\dots 1 \right\}$ over $f-1$ steps. The total number of steps remains $N$. In contrast to Baseline 2, DiffMesh demonstrates the capability to decode specific motion patterns within the same number of steps, while generating more precise and smooth output (verified in Sec. \ref{ablation_study}).


\begin{figure}[tp]
  \centering
  \includegraphics[width=1\linewidth]{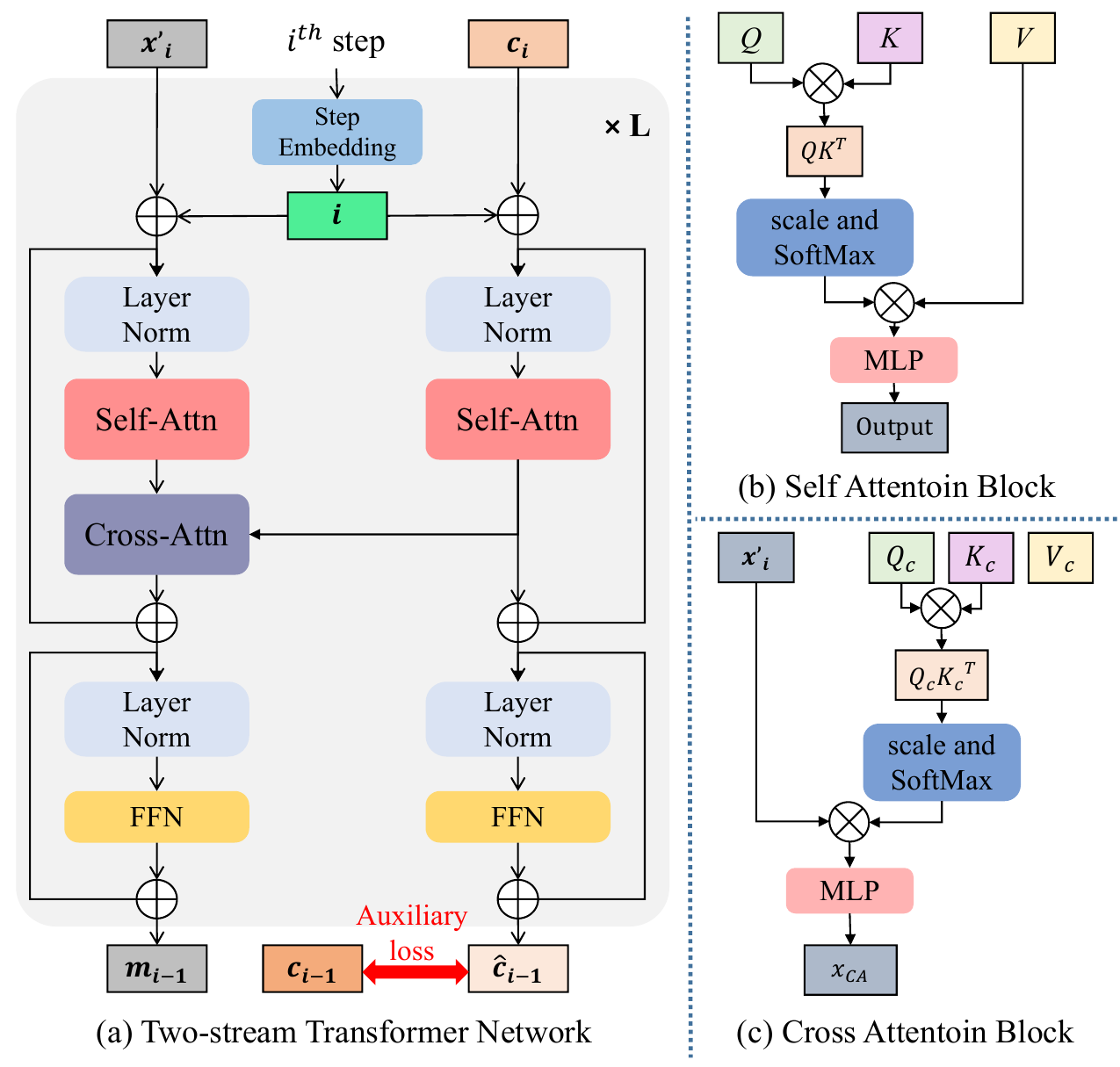}
  \vspace{-5pt}
  \caption{(a) Two-stream transformer network: Given the input feature $x'_i$ and corresponding conditional feature $c_i$, the diffusion model generates the predicted motion $m_{i-1}$ and the predicted previous conditional feature $\hat{c}_{i-1}$. (b) The self-attention block in our diffusion model. (c) The cross-attention block in our diffusion model.  The notation $\oplus$ represents element-wise addition. }
  \label{fig:md}
  \vspace{-15pt}
\end{figure}

\noindent\textbf{Network Design:} In contrast to conventional diffusion-based methods \cite{ho2020denoising,rombach2022high,peebles2023scalable} for image synthesis tasks that often rely on UNet \cite{unet} or vanilla transformer \cite{vaswani2017attention} backbones, our approach introduces a two-stream transformer design 
deeply integrated within our framework, as illustrated in Fig. \ref{fig:md}. While conventional diffusion-based methods \cite{ho2020denoising,rombach2022high,peebles2023scalable} only estimate noise during steps, our two-stream network predicts the motion features and the previous conditional feature during each step as depicted in Fig. \ref{fig:md} (c). 
This architecture 
enhances information integration through a two-stream process. Initially, it captures mesh and condition dependencies separately using two self-attention blocks. Then, these captured dependencies are merged by a cross-attention block efficiently to decode motion features. An \textbf{auxiliary loss} is computed by measuring the difference (MSE loss) between the predicted $\hat{c}_{i-1}$ and the ground truth $c_{i-1}$, which contributes to training a better transformer-based network during the reverse process.

Given the input feature $x'_{i}$ and corresponding conditional feature $c_{i}$ at each step, our network produces the predicted motion \textcolor{orange}{$\hat{m}_{i-1}$} as well as the predicted previous conditional feature $\hat{c}_{i-1}$. The feature of initial distribution is denoted as \textcolor{orange}{$x_{end}$}. Similar to Equation \ref{eq:reverse_xt},  the mesh feature $x'_{i-1}$ in DiffMesh can be computed as 
\vspace{-5pt}
\begin{align}
\small
    x'_{i-1} = \frac{1}{\sqrt{\beta_{i}}} (x'_i - \frac{1-\beta_i}{\sqrt{1-\hat{\beta_i}}} \textcolor{orange}{\hat{m}_{i-1}} )+\sigma_t \textcolor{orange}{x_{end}},
\label{eq:diffmesh_xt}
\end{align}
\vspace{-10pt}



Through our design, we efficiently acquire the output features for each frame. In comparison to Baseline 2, we maintain a total of $N$ steps while explicitly considering motion patterns during both the forward process and the reverse process. This approach results in improved accuracy and motion smoothness in the estimated mesh sequence, achieved through successive steps across frames.

\subsection{Overall Training and Loss Functions}

Based on the assumption that human motion consistently impacts the human mesh across consecutive frames, we postulate the existence of a latent space where the motion pattern $m_i$ can be conceptualized as a specific noise introduced to the mesh feature of the preceding frame $x_{i}$, as depicted in Eq. \ref{eq:diffmesh_xi}. Our objective is to derive such motion patterns for each time interval within this specific latent space during the reverse steps. Ultimately, the final mesh sequence is generated based on the estimated mesh feature \(x_{i}\) through MLP layers and the SMPL model.


We supervise the estimated SMPL parameters of each frame ($\Theta_i$ and $\beta_i$) where $i \in \left\{ 1,\dots f \right\}$ with the ground-truth using the $L2$ loss during training. Following \cite{kocabas2020vibe,tcmr}, the 3D joint coordinates are regressed by forwarding the SMPL parameters to the SMPL human body model. 
Furthermore, as mentioned in Sec. \ref{DiffMesh}, Mean Squared Error (MSE) loss is applied between the conditional feature $c_{i}$ and the predicted conditional features $\hat{c_{i}}$.  
More detailed information is included in the \textcolor{blue}{supplementary  Sec. 2}.

\vspace{-5pt}
\section{Experiments}

\begin{table*}[htp]
\renewcommand\arraystretch{1.2}
\centering
  \resizebox{1\linewidth}{!}
  {
\begin{tabular}{l|c|c|cccc|cccc}
\hline
\multicolumn{1}{c|}{}                   &           &                  & \multicolumn{4}{c|}{Human3.6M} & \multicolumn{4}{c}{3DPW} \\ \hline
\multicolumn{1}{c|}{Video-based Method} & \multicolumn{1}{c|}{Venue}     & Backbone & MPVE $\downarrow$    & MPJPE $\downarrow$  & PA-MPJPE $\downarrow$  & ACC-ERR $\downarrow$  & MPVE $\downarrow$  & MPJPE $\downarrow$ & PA-MPJPE $\downarrow$  & ACC-ERR $\downarrow$ \\ \hline
\textit{\textbf{Fixed backbone}}                         &           &                  &         &         &            &       &       &          \\ \hline
VIBE \cite{kocabas2020vibe}                                    & CVPR 2020 & ResNet50\cite{resnet}               & -       & 65.6    & 41.4  & -   & -  & 91.9  & 57.6   & 25.4  \\
TCMR \cite{tcmr}                                   & CVPR 2021 & ResNet50\cite{resnet}               & -       & 62.3    & 41.1 & 5.3      & 102.9 & 86.5  & 52.7 & 6.8    \\
MPS-Net \cite{MPS-Net}                                & CVPR 2022 & ResNet50\cite{resnet}               & -       & 69.4    & 47.4 & 3.6      & 99.7  & 84.3  & 52.1   & 7.4  \\
GLoT \cite{GLoT}                                   & CVPR 2023 & ResNet50\cite{resnet}               & -       & 67.0      & 46.3 & 3.6     & 96.3  & 80.7  & 50.6 & 6.6     \\
STAF \cite{STAF}                                   & TCSVT 2024 & ResNet50\cite{resnet}               & -       & -      & - & -     & 95.3  & 80.6  & \textbf{48.0} & 8.2     \\
\rowcolor{RowColor} DiffMesh (ours)                                   &  & ResNet50\cite{resnet}               & -       & \textbf{65.3}      & \textbf{41.9} & \textbf{3.3}     & \textbf{94.0}  & \textbf{77.2}  & 48.5 & \textbf{6.3}     \\ \hline
\textit{\textbf{Other backbone}}                         &           &                  &         &         &            &       &       &          \\ \hline
MAED \cite{MAED}                                   & ICCV 2021 & ResNet50\cite{resnet}+ViT\cite{Dosovitskiy2020ViT}               & 84.1    & 60.4    & 38.3 & -      & 93.3  & 79.0  & 45.7 & 17.6     \\
MotionBERT \cite{MotionBERT2022}                             & ICCV 2023 & DSTformer \cite{MotionBERT2022}               & 65.5    & 53.8    & 34.9 & -      & 88.1  & 76.9  & 47.2  & -   \\
\rowcolor{RowColor} DiffMesh (Ours)                               &           & DSTformer \cite{MotionBERT2022}                &   \textbf{64.2}      &  \textbf{52.5}   & \textbf{33.5}   & \textbf{5.5}    & \textbf{86.4}  & \textbf{75.7}  & \textbf{45.6} & \textbf{6.1}    \\ \hline
\textit{\textbf{With Refinement}}                         &           &                  &         &         &            &       &       &          \\ \hline
DND \cite{li2022dnd} $\dagger$                        & ECCV 2022 & HybrIK \cite{li2021hybrik}               &    -     & 52.5    & 35.5   & 6.1    & 88.6  & 73.7  & 42.7  & 7.0   \\
MotionBERT \cite{MotionBERT2022}  $\dagger$                  & ICCV 2023 & DSTformer \cite{MotionBERT2022}                & 52.6    & 43.1    & 27.8 & -      & 79.4  & 68.8  & 40.6  & -   \\
\rowcolor{RowColor} DiffMesh (Ours)  $\dagger$                   &           & DSTformer \cite{MotionBERT2022}                &   \textbf{52.1}      & \textbf{42.4}    & \textbf{27.7}   &\textbf{5.7}     & \textbf{78.4}  & \textbf{67.8}  & \textbf{40.1}  & \textbf{6.3}   \\ \hline
\end{tabular}
}
\caption{Performance comparison with SOTA video-based methods on Human3.6M and 3DPW datasets. 
  The symbol “$\dagger$” denotes the HybrIK \cite{li2021hybrik} is applied for the refinement. 
  3DPW Training set is used during training. 
  }
\label{tab: alldataset}
\vspace{-15pt}
\end{table*}

\subsection{Implementation Details}
\label{imp_detail}

For fair comparisons, we adhere to the standard implementation details commonly employed in the video-based HMR task, consistent with prior works such as \cite{kocabas2020vibe, tcmr, MPS-Net, MotionBERT2022}. We set the length of the input sequence $f$ to 16. The number of steps $N=30$. We adopt the same variance schedules as DiffPose \cite{gong2023diffpose} during our training and testing. Our approach is implemented using PyTorch \cite{PyTorch} on a single NVIDIA A5000 GPU. The optimization of weights is conducted using the Adam optimizer \cite{kingma2014adam}, with an initial learning rate of $5e^{-6}$ and weight decay set to 0.98. During the training phase, we utilize a batch size of 64, and the training process spans 60 epochs. More detailed information about the conditional feature generation part is included in the \textcolor{blue}{supplementary  Sec. 2}.

\subsection{Datasets and Evaluation Metrics}

\textbf{Datasets:} Following the previous works \cite{kocabas2020vibe,tcmr,MPS-Net}, a mixed 2D and 3D datasets are used for training. PoseTrack \cite{posetrack} and InstaVariety \cite{hmmr} are 2D datasets where 2D ground-truth annotations are provided for PoseTrack \cite{posetrack}, while pseudo ground-truth 2D annotations are generated from \cite{openpose} for InstaVariety \cite{hmmr}.  3DPW \cite{pw3d2018}, Human3.6M \cite{h36m_pami}, and AMASS \cite{AMASS2019} are 3D datasets for training. 

\noindent \textbf{Evaluation Metrics:} The HMR performance is evaluated by four standard metrics: Mean Per Joint Position Error (MPJPE), Procrustes-Aligned MPJPE (PA-MPJPE), Mean Per Vertex Position Error (MPVE), and Acceleration Error (ACC-ERR). Particularly, MPJPE, PA-MPJPE, and MPVE indicate the accuracy of the estimated 3D human pose and shape measured in millimeters (mm). ACC-ERR is proposed in HMMR \cite{hmmr} for evaluating temporal smoothness, which computes the average difference between the predicted and ground-truth acceleration of each joint in ($mm/s^2$).

\subsection{Comparison with State-of-the-art Methods}
We compare our DiffMesh with previous SOTA video-based methods on Human3.6M and 3DPW datasets (results on MPI-INF-3DHP \cite{mpi3dhp2017} are provided in the \textcolor{blue}{supplementary  Sec. 3}). Among these datasets, Human3.6M is an indoor dataset while 3DPW contains complex outdoor scenes. Following \cite{tcmr,MPS-Net}, 3DPW training sets are involved in training. The results are shown in Table \ref{tab: alldataset}. \textit{Our DiffMesh outperforms all previous methods across all evaluation metrics on the Human3.6M and 3DPW datasets}. This superior performance validates the efficacy of our DiffMesh framework for recovering high-quality human mesh structures from video input. Specifically, 
Without involving an additional refinement procedure, our DiffMesh surpasses the second-best method MotionBERT \cite{MotionBERT2022} (which already exhibits significant improvements over its predecessors) by more than 1 mm of MPJPE, PA-MPJPE, and MPVE. When incorporating HybrIK \cite{li2021hybrik} as an additional refinement procedure, our DiffMesh consistently surpasses MotionBERT \cite{MotionBERT2022} with refinement, which significantly enhances the performance of video-based HMR. 

\begin{table}[htp]
\scriptsize
\renewcommand\arraystretch{1.2}
\centering
  \resizebox{1\linewidth}{!}
  {
\begin{tabular}{l|c|cccc}
\hline
\multicolumn{1}{c|}{}                   &                 & \multicolumn{4}{c}{3DPW} \\ \hline
\multicolumn{1}{c|}{Video-based Method}    & Backbone & MPVE $\downarrow$    & MPJPE $\downarrow$ & PA-MPJPE $\downarrow$  & ACC-ERR $\downarrow$ \\ \hline
VIBE \cite{kocabas2020vibe} & Fixed ResNet50\cite{resnet}  & 113.4  & 93.5  & 56.5 & 27.1  \\
TCMR \cite{tcmr}   & Fixed ResNet50\cite{resnet}   & 111.5 & 95.0  & 55.8 & 7.0   \\
MPS-Net \cite{MPS-Net}   & Fixed ResNet50\cite{resnet}     & 109.6  & 91.6  & 54.0   & 7.5  \\
GLoT \cite{GLoT}   & Fixed ResNet50\cite{resnet}        & 107.8  & 89.9  & 53.5 & 6.7     \\
\rowcolor{RowColor} DiffMesh (ours)        & Fixed ResNet50\cite{resnet}     & \textbf{105.9}  & \textbf{88.7}  & \textbf{53.0} & \textbf{6.5}     \\ \hline
\end{tabular}
}
\caption{Performance comparison with SOTA video-based methods on 3DPW without using 3DPW training set during training. 
  }
\label{tab: 3dpw-no}
\vspace{-15pt}
\end{table}

Following previous methods \cite{tcmr,MPS-Net,GLoT}, we report the results on the 3DPW \cite{pw3d2018} without using the 3DPW training set in Table. \ref{tab: 3dpw-no}. To ensure a fair comparison, we also fixed ResNet50 \cite{resnet} as our backbone. Our DiffMesh consistently surpasses those methods across all evaluation metrics.

\subsection{Ablation Study}
\label{ablation_study}
We conduct the ablation study on 3DPW \cite{pw3d2018} dataset since it provides ground-truth SMPL parameters. All implementation details are the same as in Section \ref{imp_detail}.

\begin{table}[htp]
\renewcommand\arraystretch{1.2}
\vspace{-5pt}
\centering
  \resizebox{1\linewidth}{!}
  {
\begin{tabular}{c|cc|ccc|cc}
\hline
           & \begin{tabular}[c]{@{}c@{}}\# of input \\ frames\end{tabular} & \begin{tabular}[c]{@{}c@{}}\# of output \\ frames\end{tabular}  & MPVE $\downarrow$ & MPJPE $\downarrow$ & ACC-ERR  $\downarrow$ & \cellcolor[HTML]{FFFFFF}\begin{tabular}[c]{@{}c@{}}Total processing \\ time\end{tabular} & \cellcolor[HTML]{FFFFFF}\begin{tabular}[c]{@{}c@{}}Average time \\ per frame\end{tabular} \\ \hline
\textit{\textbf{Image-based}}                         &           &                  &         &         &            &       &            \\ \hline
I2L-MeshNet \cite{Moon_I2L_MeshNet}  & 1                                                            & 1                                                         & 110.1  & 93.2  & 30.9     & 41.1 ms                                                                                  & 41.1 ms/frame                                                                                  \\ 
HybrIK \cite{hybrik}       & 1                                                            & 1                                                         & 89.7  & 76.2  & 22.8     &  38.9 ms                                                                                  & 38.9 ms/frame                                                                                  \\
POTTER \cite{zheng2023potter}       & 1                                                            & 1                                                         & 87.4  & 75.0  & 23.1    &  42.7 ms                                                                                  & 42.7 ms/frame                                                                                  \\ \hline
\textit{\textbf{Video-based}}                         &           &                  &         &         &            &       &            \\ \hline
DND \cite{li2022dnd}       & 32                                                            & 32                                                           & 88.6  & \textbf{73.7}      & \underline{7.0}     &  2697.6 ms   &   84.3 ms/frame \\
Baseline 1 & 16                                                            & 16   & \underline{87.2}             &  77.6         &   16.5      &  2168.4 ms                                                                                        & 135.5 ms/frame                                                                                              \\
Baseline 2 & 16                                                            & 16          &  89.0         &    76.5                      &   7.9      &   224.7 ms                                                                                         & 14.0 ms/frame                                                                                           \\ \hline
\rowcolor{RowColor} DiffMesh (Ours)  & 16                                                            & 16       & \textbf{86.4} & \underline{75.7}   &  \textbf{6.1}       &    223.1 ms           & \textbf{13.9 ms/frame}                                                                                            \\ \hline
\end{tabular}
}
\vspace{-5pt}
\caption{Reconstruction performance and inference time comparison on 3DPW dataset between our DiffMesh and previous video-based HMR methods with the same hardware platform. A single NVIDIA A5000 GPU is used \textbf{with a batch size of 1 (the input of [1, num of frames, 224, 224, 3])} for fair comparison. }
\label{tab:ab_fps}
\vspace{-10pt}
\end{table}

\noindent \textbf{Comparison with the Baselines:} In Table \ref{tab:ab_fps}, we present a performance comparison among Baselines (in Section \ref{Baseline}) and our DiffMesh. Baseline 1 applies the diffusion model frame by frame, achieving better MPVE and PA-MPJPE results than Baseline 2. However, it falls short in terms of ACC-ERR. When adopting the concatenation and separation strategy in Baseline 2, the ACC-ERR is improved, but with a slight downgrade in MPVE and PA-MPJPE. Neither Baseline 1 nor Baseline 2 surpass the previous method DND \cite{li2022dnd}. In contrast, our DiffMesh outperforms both Baseline 1 and Baseline 2 across all evaluation metrics.

\begin{figure}[htp]
\vspace{-10pt}
  \centering
  \includegraphics[width=1\linewidth]{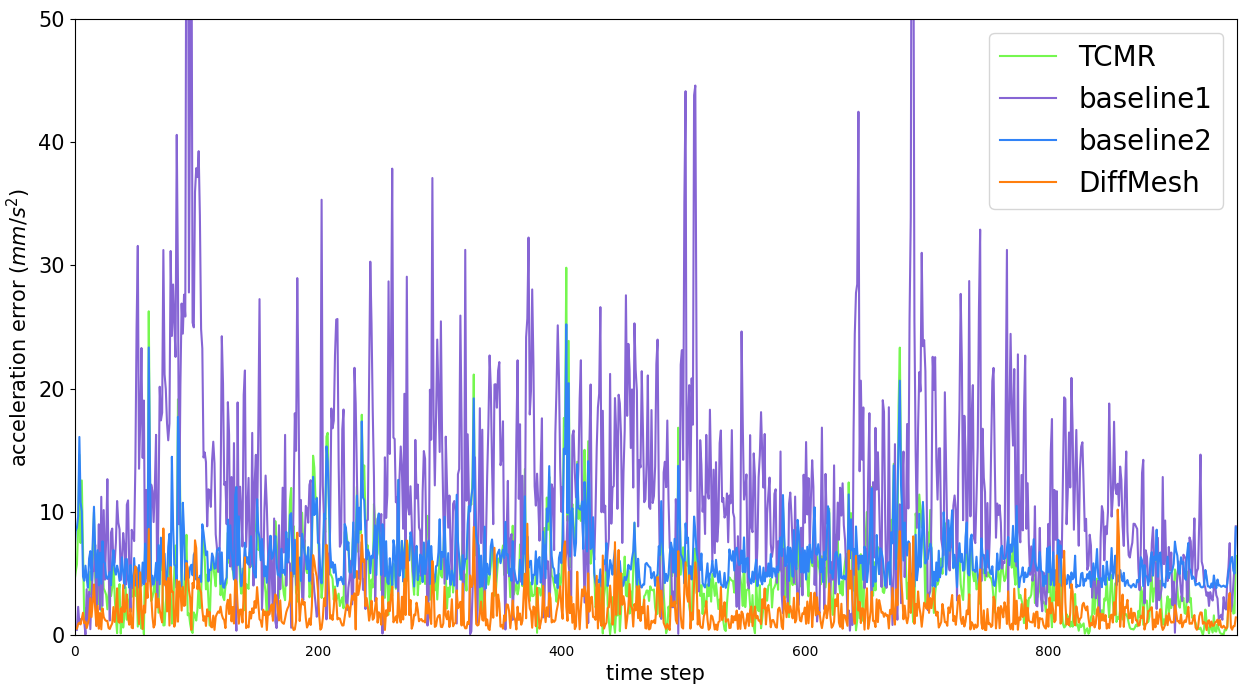}
  \vspace{-15pt}
  \caption{Acceleration error comparison of the ‘courtyard\_basketball\_01’ sequence for TCMR \cite{tcmr}, two baselines, and our DiffMesh. }
  \label{fig:ab_acc}
  \vspace{-10pt}
\end{figure}

\noindent \textbf{Temporal consistency and motion smoothness:} To further validate the enhancement in temporal consistency achieved by DiffMesh, we conducted an evaluation of acceleration error as shown in Table \ref{tab:ab_fps}. While some image-based methods can achieve impressive performance in terms of accuracy (low MPVE and MPJPE), they often suffer from motion jitter and temporal inconsistency issues. For instance, we observed that methods like HybrIK \cite{li2021hybrik} and POTTER \cite{zheng2023potter} exhibit lower MPVE and MPJPE, but concurrently, their ACC-ERR are notably higher compared to video-based HMR methods
. Instead, our DiffMesh, which integrates motion patterns into the diffusion model and enhances temporal consistency through successive steps across frames, manifests in its superior performance. DiffMesh demonstrates a 12.8\% reduction in ACC-ERR compared to the SOTA DND \cite{li2022dnd} even with fewer input frames (16 vs. 32). We further visualize the comparison of the ACC-ERR in Fig. \ref{fig:ab_acc} (``courtyard\_basketball\_01' of 3DPW dataset). The ACC-ERR for DND \cite{li2022dnd} is unavailable in Fig. \ref{fig:ab_acc} as their code has not been fully released. This highlights the superiority of DiffMesh, which leverages the consideration of human motion patterns within the diffusion model, resulting in enhanced overall performance, especially in terms of motion smoothness.

\noindent \textbf{Inference time analysis:} A prevalent consideration regarding diffusion-based methods pertains to their inference speed,  which is also a crucial evaluation metric besides accuracy. 
Our Baseline 1 requires a long processing time to output high-quality human mesh. Our Baseline 2 reduced the processing time but the acceleration error is still worse than DND \cite{li2022dnd}.  Compared to these methods, 
DiffMesh not only achieves the best mesh recovery performance and motion smoothness, while maintaining a competitive processing time. The average time can achieve 13.9 ms/frame, which is much faster than previous SOTA methods (More detailed comparisons with previous methods are provided in the \textcolor{blue}{supplementary  Sec. 3} when considering some methods can extract all features by their backbone and then utilize batch processing to accelerate the inference speed). 
This combination of accuracy and efficiency makes DiffMesh a promising diffusion solution for video-based human mesh recovery applications.

\begin{figure}[htp]
\vspace{-5pt}
  \centering
  \includegraphics[width=0.98\linewidth]{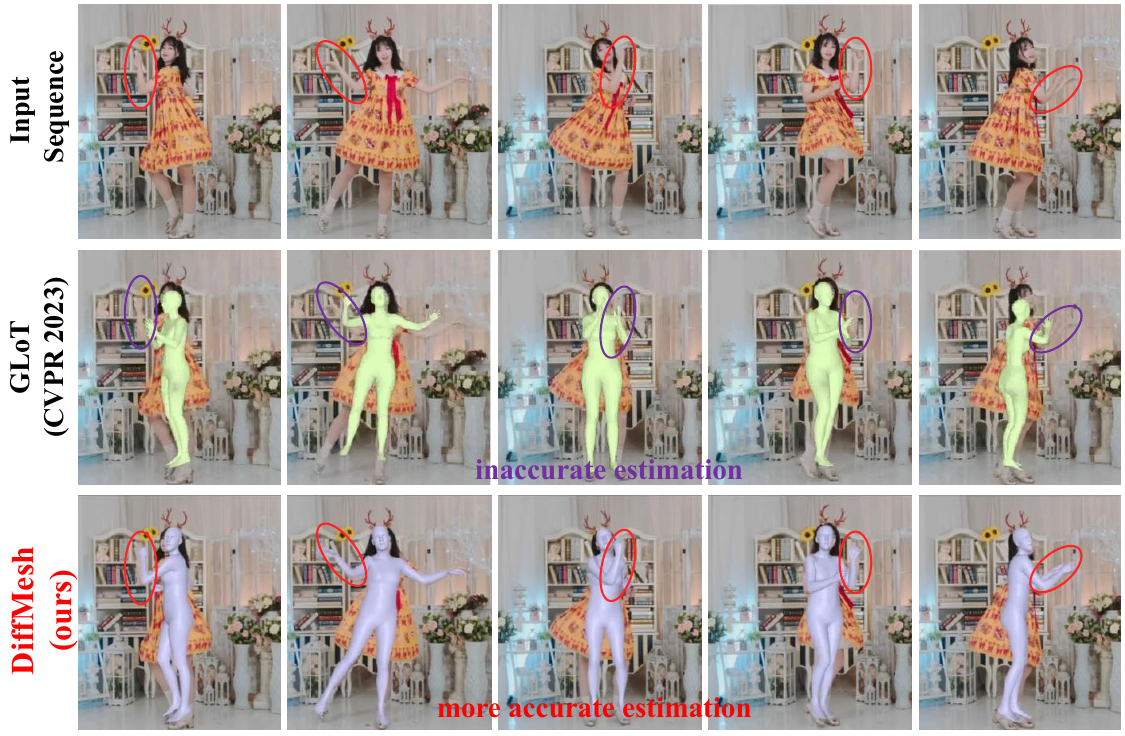}
  \vspace{-5pt}
  \caption{The in-the-wild visual comparison between recent GLoT \cite{GLoT} with our DiffMesh. The circles highlight locations where DiffMesh is more accurate than GLoT. More examples are provided in the \textcolor{blue}{supplementary  Sec. 4 and in demo videos}. }
  \label{fig:ab_vis}
  \vspace{-10pt}
\end{figure}

\subsection{In-the-Wild Visualization}
To demonstrate the practical performance of DiffMesh, we present the qualitative comparison between DiffMesh and a recent GLoT \cite{GLoT} on in-the-wild video in Fig. \ref{fig:ab_vis}. Although achieving good performance in terms of mesh recovery performance on common datasets, GLoT \cite{GLoT} suffers from temporal inconsistency when applied for in-the-wild video. 
As shown in Fig. \ref{fig:ab_vis}, GLoT \cite{GLoT} fails to generate the accurate mesh. In contrast, our DiffMesh enhances motion smoothness and reconstruction accuracy through our proposed motion-aware diffusion framework. 

\section{Conclusion}
In this paper, we present a novel diffusion framework (DiffMesh) for human mesh recovery from a video. DiffMesh innovatively connects diffusion models with human motion, resulting in the efficient generation of highly precise and seamlessly smooth output mesh sequences. Extensive experiments are conducted on Human3.6M and 3DPW datasets showing impressive performance. 

While DiffMesh demonstrates its superior performance for in-the-wild video input, like previous methods, it may produce unrealistic mesh outputs in scenarios with significant occlusions. Our future research will focus on exploring spatial-temporal interactions within the human body to mitigate this occlusion challenge.

\newpage
\appendix

\noindent \textbf{\large{Supplementary Material}}

\section{Overview}

The supplementary material is organized into the following sections:

\begin{itemize}
\item Section \ref{detail}: More related Work and implementation details.

\item Section \ref{exp}: Mathematical proof and more experiments about the number of input frames, the two-stream transformer network, initial distributions, and the auxiliary loss. 

\item Section \ref{meshvis}: More human mesh visualization.

\item Section \ref{Broader}: Broader impact and limitation
\end{itemize}




\section{More Related Work and Implementation Details}
\label{detail}
\subsection{Related Work}
The majority of methods  ~\cite{kanazawaHMR18,Kolotouros2019SPIN,Choi_2020_ECCV_Pose2Mesh,gtrs,zheng2023feater,you2023gator,zheng2023potter,li2023niki} for HMR rely on a parametric human model, such as SMPL \cite{SMPL:2015}, to reconstruct the mesh by estimating pose and shape parameters. As a fundamental HMR work, SPIN ~\cite{Kolotouros2019SPIN} combines regression and optimization in a loop, where the regressed output serves as better initialization for optimization (SMPLify). METRO \cite{lin2021metro} is the first transformer-based method that models vertex-vertex and vertex-joint interaction using a transformer encoder after extracting image features with a CNN backbone. HybrIK \cite{li2021hybrik} and HybrIK-X \cite{li2023hybrik} present novel hybrid inverse kinematics approaches that transform 3D joints to body-part rotations via twist-and-swing decomposition. Lin et al. \cite{lin2023one} propose a one-stage pipeline for 3D whole-body (body, hands, and face) mesh recovery. PyMAF \cite{pymaf2021} and its extension work PyMAF-X \cite{zhang2023pymafx} capitalize on a feature pyramid to rectify predicted parameters by aligning meshes with images, extracting mesh-aligned evidence from finer-resolution features. CLIFF \cite{li2022cliff} enhances holistic feature representation by incorporating bounding box information into cropped-image features. It employs a 2D reprojection loss considering the full frame and leverages global-location aware supervision to directly predict global rotation and more accurately articulated poses. ReFit \cite{wang2023refit} proposes a feedback-update loop reminiscent of solving inverse problems via optimization, iteratively reprojections keypoints from the human model to feature maps for feedback, and utilizes a recurrent-based updater to refine the model's fit to the image. HMR2.0 \cite{goel2023humans} develops a system that can simultaneously reconstruct and track humans from video, but only reports the frame-based results for the HMR task without considering temporal information. Foo et al. \cite{foo2023distribution} first introduce a diffusion-based approach for recovering human mesh from a single image. The recovered human mesh is obtained by the reverse diffusion process. However, when applied to video sequences, these image-based methods suffer from severe motion jitter due to frame-by-frame reconstruction, making them unsuitable for practical use. 

Compared to image-based HMR methods, video-based methods \cite{zeng2022deciwatch,smoothnet,you2023coeval} utilize temporal information to enhance motion smoothness from video input. In addition to the methods \cite{kocabas2020vibe,AMASS2019,meva,tcmr,MPS-Net,MotionBERT2022} introduced in the main paper, there are several other noteworthy approaches for video-based HMR.  Kanazawa et al.\cite{hmmr} first propose a convolutional network to learn human motion kinematics by predicting past, current, and future frames. Based on \cite{hmmr}, Sun et al. \cite{sun2019human} further propose a self-attention-based temporal model to improve performance. DND \cite{li2022dnd} utilizes inertial forces control as a physical constraint to reconstruct 3D human motion. GLoT \cite{GLoT} adopts a novel approach by decoupling the modeling of short-term and long-term dependencies using a global-to-local transformer. PMCE \cite{you2023coeval} follows a two-step process, where it first estimates 3D human pose and then regresses the mesh vertices through a co-evaluation decoder that takes into account the interactions between pose and mesh.

\subsection{Datasets}
\noindent \textbf{3DPW} \cite{pw3d2018} is a dataset that captures outdoor and in-the-wild scenes using a hand-held camera and a set of inertial measurement unit (IMU) sensors attached to body limbs. The ground-truth SMPL parameters are computed based on the returned values. This dataset includes 60 videos of varying lengths, and we use the official split to train and test the model. The split comprises 24, 12, and 24 videos for the training, validation, and test sets, respectively. The MPJPE, PA-MPJPE, MPJVE, and ACC-ERR are reported when evaluating this dataset.

\noindent \textbf{Human3.6M} \cite{h36m_pami} is a large-scale benchmark for the indoor 3D human pose. It includes 15 action categories and 3.6M video frames. Following \cite{kocabas2020vibe,tcmr,MPS-Net}, we use five subjects (S1, S5, S6, S7, S8) for the training set and two subjects (S9, S11) for the testing set. The dataset is subsampled from its original 50 fps to 25 fps for both training and evaluation purposes. When calculating MPJPE and PA-MPJPE, only 14 joints are selected for a fair comparison to the previous works.

\noindent \textbf{MPI-INF-3DHP} \cite{mpi3dhp2017} is a 3D benchmark that consists of both indoor and outdoor environments. The training set includes 8 subjects, with each subject having 16 videos, resulting in a total of 1.3M video frames captured at 25 fps. The markerless motion capture system is used for providing 3D human pose annotations. The test set comprises 6 subjects performing 7 actions in both indoor and outdoor environments. Following \cite{kocabas2020vibe,tcmr,MPS-Net}, the MPJPE and PA-MPJPE are measured on valid frames, which include approximately every 10th frame, using 17 joints defined by MPI-INF3DHP. The ACC-ERR is computed using all frames.

\noindent \textbf{InstaVariety} \cite{hmmr} is a 2D human dataset curated by HMMR \cite{hmmr} , comprising videos collected from Instagram using 84 motion-related hashtags. The dataset contains 28K videos with an average length of 6 seconds, and pseudo-ground truth 2D pose annotations are acquired using OpenPose\cite{openpose}.

\noindent \textbf{PoseTrack} \cite{posetrack} is a 2D benchmark designed for multi-person pose estimation and tracking in videos. This dataset comprises 1.3K videos and 46K annotated frames, captured at varying fps around 25 fps. There are 792 videos used for the official train set, which includes 2D pose annotations for 30 frames located in the middle of each video.

\subsection{Loss Function}

Our DiffMesh relies on the SMPL model \cite{SMPL:2015} to reconstruct the human mesh. The SMPL model can generate the body mesh $M \in \mathbb{R}^{N \times 3} $ with $N=6890$ vertices by taking in the predicted pose parameters $\theta$ and the shape parameters $\beta$ as inputs, which can be expressed as $M = SMPL (\theta, \beta) $. Once the body mesh $M$ is obtained, the body joints $J$ can be estimated by applying the predefined joint regression matrix $W$, i.e., $J \in \mathbb{R}^{k \times 3} = W \cdot M$, where $k$ represents the number of joints. We adopt the same loss function as previous methods TCMR \cite{tcmr}.

\begin{equation}
\begin{aligned} 
  \begin{aligned}
\mathcal{L}_{HMR} &=  w_1 \| \beta - \beta^*  \| + w_2 \| \theta - \theta^*  \| + w_3 \| J - J^*  \|\\
  \end{aligned}\\
\end{aligned}
\end{equation}
\noindent where * denote the ground-truth value,  $w_1=0.5$, $w_2=10$, and $w_3=1000$.

Besides this general loss for mesh recovery, we add additional auxiliary loss as mentioned in Section 3.4 of the main paper. Our designed transformer-based diffusion model can predict the previous conditional feature $\hat{c}_{i-1}$ given the current conditional feature input  $c_{i}$. A MSE loss is applied between the ground truth $c_{i-1}$ and predicted $\hat{c}_{i-1}$:
\begin{equation}
\begin{aligned} 
  \begin{aligned}
\mathcal{L}_{aux} &=  \| c_{i-1} - \hat{c}_{i-1} \|_2^2\\
  \end{aligned}\\
\end{aligned}
\end{equation} 
This auxiliary loss contributes to the refinement of our transformer-based diffusion model during the training process. Thus, the overall loss for our DiffMesh is the sum of the $\mathcal{L}_{HMR} $ and $\mathcal{L}_{aux}$:
\begin{equation}
\begin{aligned} 
  \begin{aligned}
\mathcal{L}_{overall} &= \mathcal{L}_{HMR} +   w_4 \mathcal{L}_{aux} \\
  \end{aligned}\\
\end{aligned}
\end{equation} 
\noindent where $w_4=0.01$.



\subsection{More Details about the Architecture}

\begin{figure}[tp]
\vspace{-5pt}
  \centering
  \includegraphics[width=1\linewidth]{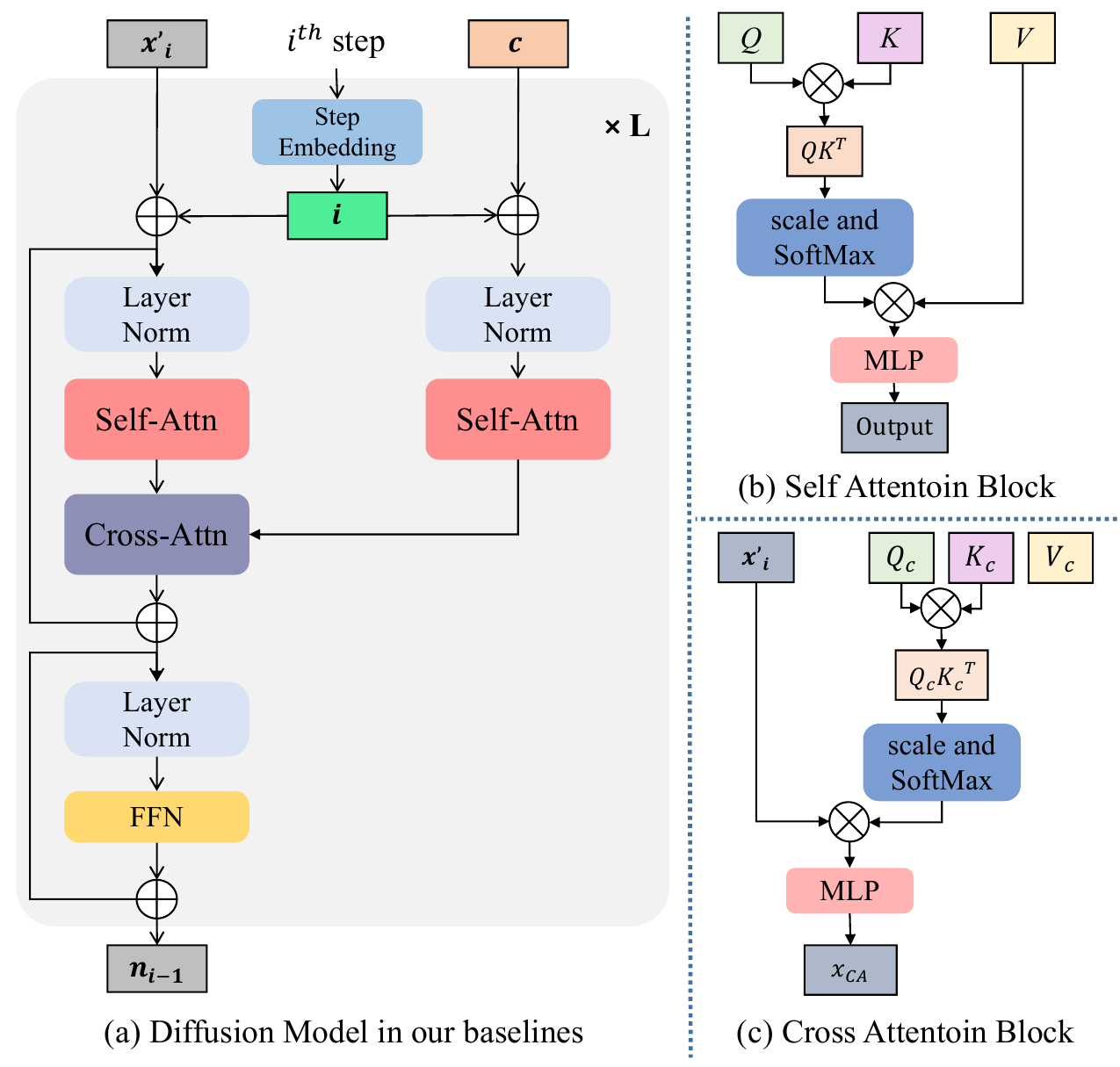}
  \vspace{-20pt}
  \caption{The diffusion model in our baselines}
  \label{fig:supp_diffmodel}
  \vspace{-10pt}
\end{figure}  

\textbf{Diffusion model in our baselines:}
The architecture of the diffusion model employed in our baselines is illustrated in Fig. \ref{fig:supp_diffmodel}. It shares similarities with the architecture within our DiffMesh, featuring two self-attention blocks designed to capture global dependencies and one cross-attention block focused on integrating information between the denoising input $x_{i}$ and the constant conditional feature $c$. In the baseline approach, as the conditional feature $c$ remains the same throughout the denoising process, there is no need to estimate the conditional feature for each subsequent denoising step. Thus, it only return the estimated noise term $n_{i-1}$.

\begin{figure*}[htp]
\vspace{-5pt}
  \centering
  \includegraphics[width=1\linewidth]{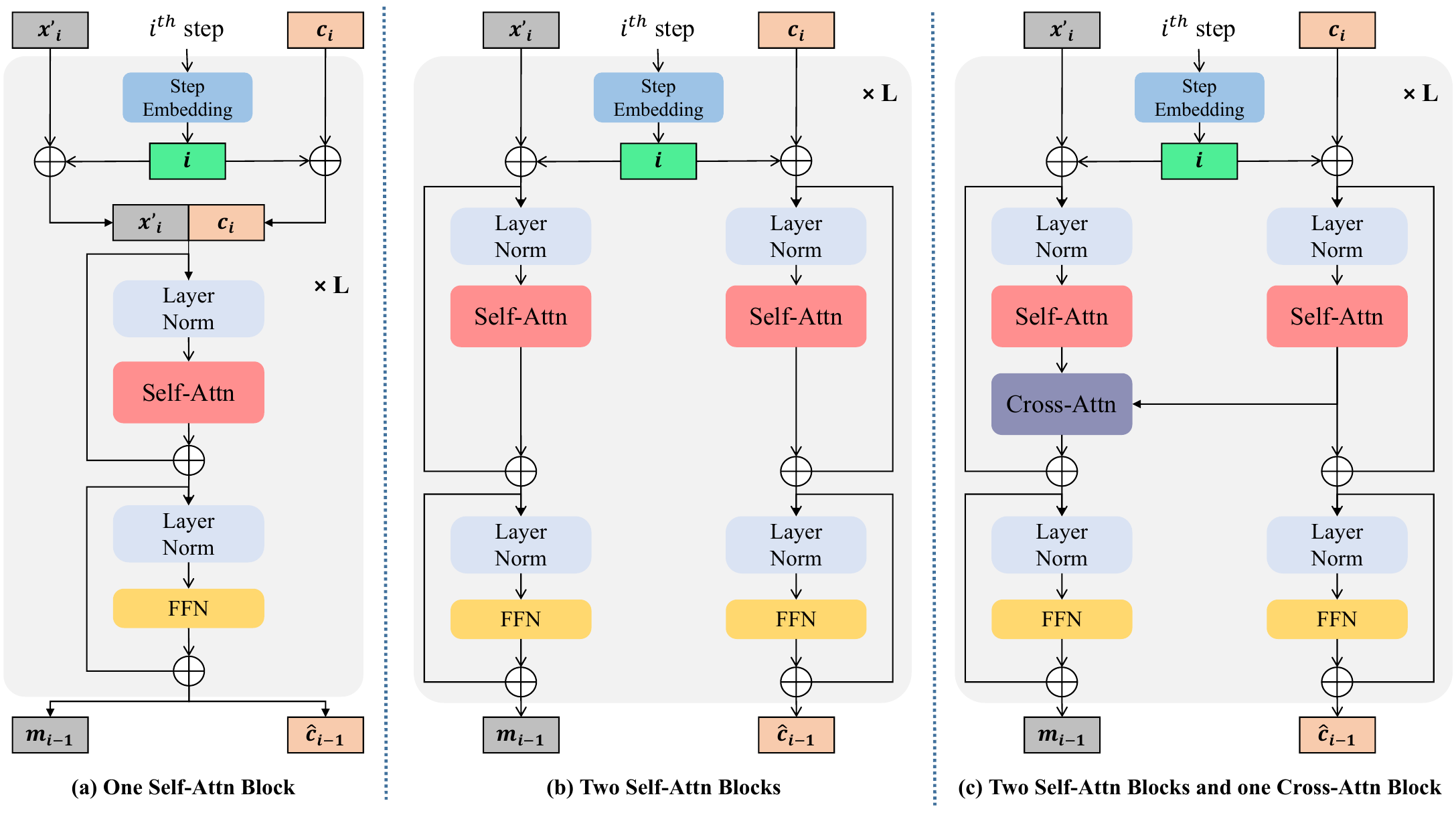}
  \vspace{-15pt}
  \caption{Different design choices of the transformer architecture: (a) Only one Self-Attn Block. (b) Two Self-Attn Blocks. (c) Two Self-Attn Blocks and one Cross-Attn Block.}
  \label{fig:supp_transformer}
  \vspace{-5pt}
\end{figure*}

\textbf{Conditional features generation block:}
Our chosen backbone to extract features for both our proposed method and the baselines is ResNet-50 \cite{resnet} or DSTformer \cite{MotionBERT2022}. After extracting features from each frame $b_{i}$, where $ i \in \left\{ 1,\dots f \right\}$, using the backbone, our goal is to generate $N$ conditional features to be utilized during the reverse process. To achieve this, we pad additional $N-f$ zero features, $b_{f+1}, \cdots, b_{N}$.  Then, we combine them with the existing features, creating new features $b \in \mathbb{R} ^{N \times D}$, where $D$ represents the embedded dimension. Subsequently, we apply a transformer block \cite{Dosovitskiy2020ViT} to model these features and return the required conditional features denoted as $c \in \mathbb{R} ^{N \times D}$.

\section{Mathematical Proof and More Experiments}
\label{exp}

\subsection{Mathematical Proof of modeling human motion as noise in diffusion model}

Our approach draws an analogy between human motion and noise, treating the motion between adjacent frames as a structured form of noise. By operating in a high-dimensional latent space, we capture the complexity of human motion, where small perturbations (or "noise") in this space can be modeled as Gaussian. This allows us to align the problem with the core principles of diffusion models.

Following Equation 6 in the main paper, we have

\begin{equation}
\scalebox{0.85}{$
\mathbb{E}_{x_0 \sim q} \left[ -\log p(x_0) \right] = \mathbb{E}_{x_0 \sim q} \left[ \log \mathbb{E}_{x_{1:T}, y_{0:T} \sim q} \frac{q(x_{1:T}, y_{0:T} | x_0)}{p(x_{0:T}, y_{0:T})} \right]
$}
\label{eq:supp1}
\end{equation}

Using Jensen's Inequality, we can bound Eq \ref{eq:supp1} by moving the expectation inside the logarithm:
\begin{equation}
\scalebox{0.85}{$
\mathbb{E}_{x_0 \sim q} \left[ -\log p(x_0) \right] \leq \mathbb{E}_{x_0, x_{1:T}, y_{0:T} \sim q} \left[ \log \frac{q(x_{1:T}, y_{0:T} | x_0)}{p(x_{0:T}, y_{0:T})} \right]
$}
\end{equation}

The forward process in DiffMesh is designed to model the human motion between adjacent frames as structured noise. This noise is Gaussian in the latent space, where the human motion patterns are simplified. We define the forward process for the latent motion as:
\begin{equation}
q(m_{t+1} | m_t) = \mathcal{N}(m_{t+1}; \sqrt{1 - \beta_t} m_t, \beta_t I)
\end{equation}

Here $m_t$ represents the motion noise at time step $t$, and $\beta_t$ controls the level of noise added to the motion between consecutive frames. This equation models the motion between frames as Gaussian perturbations in latent space.

We now need to decompose the joint probabilities \( q(x_{1:T}, y_{0:T} | x_0) \) and \( p(x_{0:T}, y_{0:T}) \) into their respective transition probabilities over time \( t \):

The forward process \( q(x_{1:T}, y_{0:T} | x_0) \) can be written as:
\begin{equation}
\scalebox{0.85}{$
q(x_{1:T}, y_{0:T} | x_0) = q(x_T | x_0) \prod_{t=2}^{T} q(x_{t-1} | x_t, x_0) \prod_{t=0}^{T} q(y_t | x_t)
$}
\end{equation}

Similarly, the reverse process \( p(x_{0:T}, y_{0:T}) \) is:
\begin{equation}
\scalebox{0.85}{$
p(x_{0:T}, y_{0:T}) = p(x_T) \prod_{t=T}^{1} p(x_{t-1} | x_t) \prod_{t=T}^{0} p(y_t | x_t)
$}
\end{equation}

Next, we focus on deriving the bound based on KL divergence. Substituting the decomposed expressions, we obtain two main terms: one for the states $x$ and another for the observations $y$

\begin{align}
\mathbb{E}_{x_0, x_{1:T}, y_{0:T} \sim q} \left[ \log \frac{q(x_T | x_0) \prod_{t=2}^{T} q(x_{t-1} | x_t, x_0)}{p(x_T) \prod_{t=T}^{1} p(x_{t-1} | x_t)} \right] \nonumber \\
+ \mathbb{E}_{x_0, x_{1:T}, y_{0:T} \sim q} \left[ \log \frac{\prod_{t=0}^{T} q(y_t | x_t)}{\prod_{t=T}^{0} p(y_t | x_t)} \right]
\end{align}

Each of these terms corresponds to the forward process (adding Gaussian motion noise) and the reverse process (denoising to recover the original human motion).

The final step is to express the result as a sum of KL divergences. For the forward and reverse processes of both the states $x$ and the observations $y$, we can represent this as:

\begin{align}
= D_{\text{KL}}(q(x_T | x_0) || p(x_T)) 
+ \mathbb{E}_q \left[ -\log p(x_0 | x_1) \right] \nonumber \\
+ \sum_{t=2}^{T} D_{\text{KL}}(q(x_{t-1} | x_t, x_0) || p(x_{t-1} | x_t)) \nonumber \\
+ \sum_{t=0}^{T} D_{\text{KL}}(q(y_t | x_t) || p(y_t | x_t))
\end{align}

\noindent Here \( D_{\text{KL}}(q(x_T | x_0) || p(x_T)) \) measures the divergence between the final denoised human mesh and the actual motion, and \( D_{\text{KL}}(q(x_{t-1} | x_t, x_0) || p(x_{t-1} | x_t)) \) measures the divergence at each intermediate step in recovering the motion, helping to maintain temporal consistency. Thus, we can derive Equation 7 in the main paper. 

\subsection{Performance on MPI-INF-3DHP dataset}

\begin{table}[htp]
    \centering
    \vspace{-5pt}
    \resizebox{1\linewidth}{!}
  {
    \begin{tabular}{c|ccc}
\hline
\multicolumn{1}{l|}{} & \multicolumn{3}{c}{MPI-INF-3DHP} \\ \hline
Methods               & MPJPE  $\downarrow$   & PA-MPJPE  $\downarrow$   & ACC-ERR  $\downarrow$   \\ \hline
VIBE \cite{kocabas2020vibe}                 & 103.9   & 68.9       & 27.3      \\
TCMR \cite{tcmr}                  & 97.6    & 63.5       & 8.5       \\
MAED \cite{MAED}                & 83.6    & 56.2       & -         \\
MPS-Net \cite{MPS-Net}              & 96.7    & 62.8       & 9.6       \\
GLoT  \cite{GLoT}                & 93.9    & 61.5       & 7.9       \\ 
\hline
DiffMesh (ours)       & \textbf{78.9}    & \textbf{54.4}       & \textbf{7.0}       \\ \hline
\end{tabular}}
\vspace{-5pt}
    \caption{Performance comparison with state-of-the-art methods on MPI-INF-3DHP dataset. All methods use pre-trained ResNet-50 \cite{resnet} (fixed weights) to extract features except MAED. }
    \label{tab:supp-mpii3d}
\vspace{-5pt}
\end{table} 

To conduct experiments on the MPI-INF-3DHP \cite{mpi3dhp2017} dataset, we follow the same setting as VIBE \cite{kocabas2020vibe}, TCMR \cite{tcmr}, and MPS-Net \cite{MPS-Net}
. The input features of each frame are extracted from ResNet-50 \cite{resnet} without fine-tuning for fair comparisons. The results are shown in Fig. \ref{tab:supp-mpii3d}. Our DiffMesh consistently outperforms previous methods with significant improvement (more than 5.9 mm $\downarrow$ of MPJPE, 1.8  $\downarrow$ of PA-MPJPE, and 0.7 $\downarrow$ of ACC-ERR). This showcases the remarkable performance enhancement achieved by our approach, highlighting its potential as a state-of-the-art solution for video-based human mesh recovery across various datasets and real-world applications.


\subsection{Effectiveness of the number of input frames and additional steps}
Following the same setting as previous video-based methods such as VIBE \cite{kocabas2020vibe}, TCMR \cite{tcmr}, and MPS-Net \cite{MPS-Net}, the number of input frames $f$ is set to be 16. To further investigate the impact of the number of input frames, we conduct experiments on the 3DPW dataset given the different number of input frames. The results are shown in Table. \ref{tab:supp_frame}.

In general, the performance can be improved (lower MPJPE, PA-MPJPE, MPVPE, and ACC-ERR) when the number of input frames $f$ is increased. Specifically, when maintaining the total number of steps $N$ at 30 and varying $f$ from 8 to 16 to 24, the improvements are notable. 
In our ablation study, the lowest MPVE, MPJPE, and ACC-ERR are achieved when $f=32$ with total steps of 40.  

To strike an optimal balance between efficiency and performance, it's crucial to seek improved results with a reduced total number of steps $N$. For instance, when $f=16$, the optimal $N$ is determined to be 30, demonstrating comparable results to $N=40$ at a faster processing speed. Similarly, for $f=24$, the optimal $N$ is identified as 30 based on the results.

\begin{table}[htp]
\scriptsize
\vspace{-5pt}
\centering
  \resizebox{1\linewidth}{!}
  {
\begin{tabular}{c|cc|c|ccc}
\hline
input frames & \multicolumn{1}{c}{\begin{tabular}[c]{@{}c@{}}steps for output \\ sequence\end{tabular}} & additional steps & Total steps & MPVE $\downarrow$ & MPJPE $\downarrow$ & ACC-ERR $\downarrow$ \\ \hline
8     & 7      & 0               & 7          &  89.8    &  77.9     &   6.9      \\
16     & 15       & 0               & 15           &  88.5    &  77.4     &   6.5  \\
24      & 23      & 0               & 23          & 87.6     & 75.2      &  6.2       \\\hline
8      & 7       & 13               & 20          & 88.6     & 76.9      &   6.7      \\
16     & 15       & 5               & 20          &  88.0    &  77.1     &  6.5       \\\hline
8     & 7        & 23               & 30          &  87.4    & 76.5      &   6.5      \\
16     & 15       & 15               & 30          &  86.4    & 75.7      &   6.1      \\
24     & 23       & 7               & 30          &  86.2   &   74.7     &   5.9      \\ \hline
16     & 15       & 25               & 40          &  87.1    & 75.6      &   6.2      \\
24     & 23       & 17               & 40          &  86.5    &  74.7     &  6.1       \\
32     & 31       & 8               & 40          &  86.0    &  74.9     &  5.8       \\ \hline
\end{tabular}
}
  \caption{Performance of the different number of input frames and the number of additional steps on the 3DPW dataset.}
\label{tab:supp_frame}
\vspace{-5pt}
\end{table}



\subsection{Different design choices of our transformer-based diffusion model}
As introduced in Section 3.3 of the main paper, our proposed transformer-based diffusion model consists of two self-attn blocks with one cross-attn block (also depicted in Fig. \ref{fig:supp_transformer} (c)). Given the input feature $x'_{i}$ and corresponding conditional feature $c_{i}$, the transformer-based diffusion model produces the predicted noise \textcolor{orange}{$m_{i-1}$} and the predicted previous conditional feature $\hat{c}_{i-1}$. We apply two self-attention blocks for $x'_{i}$ and  $c_{i}$ separately, then a cross-attention block is adopted to fuse the conditional features with mesh features. To validate the effectiveness, we compare this design with (a): a self-attention block applied for the concatenated features; and (b) two two self-attention blocks for $x'_{i}$ and  $c_{i}$ separately without cross-attention block. The results are shown in Table \ref{tab:supp_transformer}. Clearly, our design (c) in DiffMesh outperforms (a) and (b) for all evaluation metrics on the 3DPW dataset due to enhanced information integration using two-stream and cross-attention fusion design.

\begin{table}[htp]
\vspace{-5pt}
\centering
  \vspace{-5pt}
  \resizebox{1\linewidth}{!}
  {
\begin{tabular}{c|cccc}
\hline
                             & \multicolumn{4}{c}{3DPW}      \\ \hline
                             & MPVE $\downarrow$ & MPJPE $\downarrow$ & PA-MPJPE $\downarrow$ & ACC $\downarrow$ \\ \hline
(a) one self-attn            &  86.9    &  76.9     &  47.5        &  6.3   \\ \hline
(b) two self-attn            &  87.4    &  76.4     &  45.9        &  6.2   \\ \hline
(c) self-attn and cross attn &  \textbf{86.4}     & \textbf{75.7}  & \textbf{45.6} & \textbf{6.1}    \\ \hline
\end{tabular}
}
  \caption{Ablation study of transformer block design on 3DPW dataset.}
\label{tab:supp_transformer}
\vspace{-5pt}
\end{table}

\subsection{Effectiveness of the auxiliary loss:} 

To validate the effectiveness of our proposed auxiliary loss, we compare the results as shown in Table \ref{tab:supp_loss}, which demonstrated that our proposed auxiliary loss can help to improve the reconstruction performance (MPJPE, PA-MPJPE, and MPJVE) and the motion smoothness (ACC-ERR). 
\begin{table}[htp]
\scriptsize
\renewcommand\arraystretch{1.2}
\vspace{-10pt}
\centering
  \resizebox{1\linewidth}{!}
  {
\begin{tabular}{c|cccc}
\hline
     & \multicolumn{4}{c}{3DPW}         \\ \hline
loss & MPVPE$\downarrow$ & MPJPE$\downarrow$ & PA-MPJPE$\downarrow$ & Accel$\downarrow$ \\ \hline
Without $\mathcal{L}_{aux}$    & 86.8     & 76.0  & 47.1 & 6.2   \\ \hline
With $\mathcal{L}_{aux}$   & \textbf{86.4}     & \textbf{75.7}  & \textbf{45.6} & \textbf{6.1}   \\ \hline
\end{tabular}
}
\caption{Evaluation of the combinations of loss functions on the 3DPW dataset.}
\label{tab:supp_loss}
\vspace{-10pt}
\end{table}

\subsection{Inference Time Analysis:}

Methods like MPS-Net \cite{MPS-Net} and GLoT \cite{GLoT} only estimate the human mesh of the center frame given 16 frames as their input. Considering these methods can extract all features by their backbone once and then utilize batch processing to accelerate the inference speed, we provide a more thorough inference time comparison in Table \ref{tab:supp_fps}. 

In this experiment, the video input comprises a total of 64 frames. Upon feature extraction from the backbone (with the shape of $[64,2048]$), MPS-Net and GLoT require the creation of 64 batch input tubes $[64, 16, 2048]$ through padding and sliding window. Since their models only return the output mesh of the center frame, the output would be $[64,1, 6890,3]$, indicating output mesh vertices $[6890,3]$ across 64 frames. In contrast, our DiffMesh just needs to reshape the input $[64,2048]$ into 4 batches, resulting in the shape of $[4,16,2048]$. Consequently, the output of DiffMesh is $[4,16,6890,3]$, which is then reshaped back to $[64,6890,3]$. Based on the total processing time, our DiffMesh is more efficient than MPS-Net \cite{MPS-Net} and GLoT \cite{GLoT} since DiffMesh can output human meshes of all input frames.

\begin{table*}[htp]
\renewcommand\arraystretch{1.1}
\vspace{-5pt}
\centering
  \resizebox{1\linewidth}{!}
  {
\begin{tabular}{c|c|c|c|c|c}
\hline
\begin{tabular}[c]{@{}c@{}}Video-based\\ Methods\end{tabular} & \begin{tabular}[c]{@{}c@{}}total frames for\\  video input\end{tabular} & Backbone & \begin{tabular}[c]{@{}c@{}}features after backbone are \\ reshaped for model processing\end{tabular} &  \begin{tabular}[c]{@{}c@{}}Output\\ shape \end{tabular} & \begin{tabular}[c]{@{}c@{}}processing time\\ (without backbone time)\end{tabular} \\ \hline
MPS-Net \cite{MPS-Net}                                                      & 64                                                                      & ResNet50 \cite{resnet} & {[}64,2048{]} to {[}64,16,2048{]}   & {[}64,1,6890,3{]} to {[}64,6890,3{]}                                                                 & 1.04 s                                                                            \\
GLoT \cite{GLoT}                                                         & 64                                                                      & ResNet50 \cite{resnet} & {[}64,2048{]} to {[}64,16,2048{]}    & {[}64,1,6890,3{]} to {[}64,6890,3{]}                                                                  & 1.17 s                                                                            \\ \hline
DiffMesh (ours)                                               & 64                                                                      & ResNet50 \cite{resnet} & {[}64,2048{]} to {[}4,16,2048{]}      & {[}4,16,6890,3{]} to {[}64,6890,3{]}                                                                 & 0.34 s                                                                            \\ \hline
\end{tabular}
}
\vspace{5pt}
\caption{Inference time comparison on 3DPW dataset between our DiffMesh and previous video-based HMR methods with the same hardware platform ( single NVIDIA A5000 GPU is used).}
\label{tab:supp_fps}
\vspace{-10pt}
\end{table*}



\section{Human Mesh Visualization}
\label{meshvis}

\begin{figure}[htp]
  \centering
  \includegraphics[width=1\linewidth]{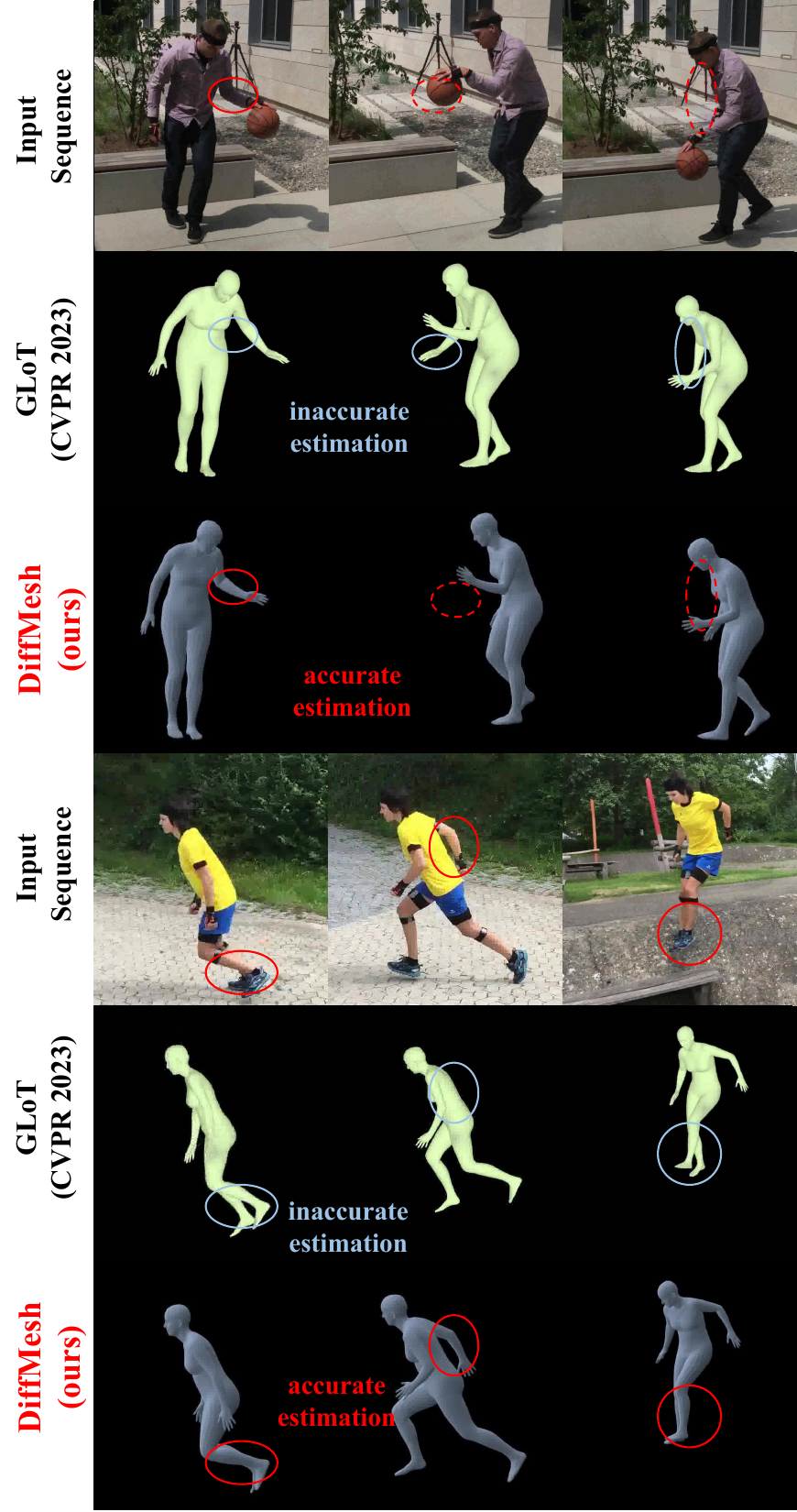}
  \vspace{-5pt}
  \caption{Qualitative comparison on the 3DPW dataset}
  \label{fig:supp_vis0}
  \vspace{-5pt}
\end{figure}

We first visualize the qualitative comparison on the 3DPW \cite{pw3d2018} dataset in Fig.~\ref{fig:supp_vis0}. The circle areas highlight locations where our DiffMesh performs better than GLoT \cite{GLoT}.

\begin{figure*}[htp]
\vspace{-5pt}
  \centering
  \includegraphics[width=1\linewidth]{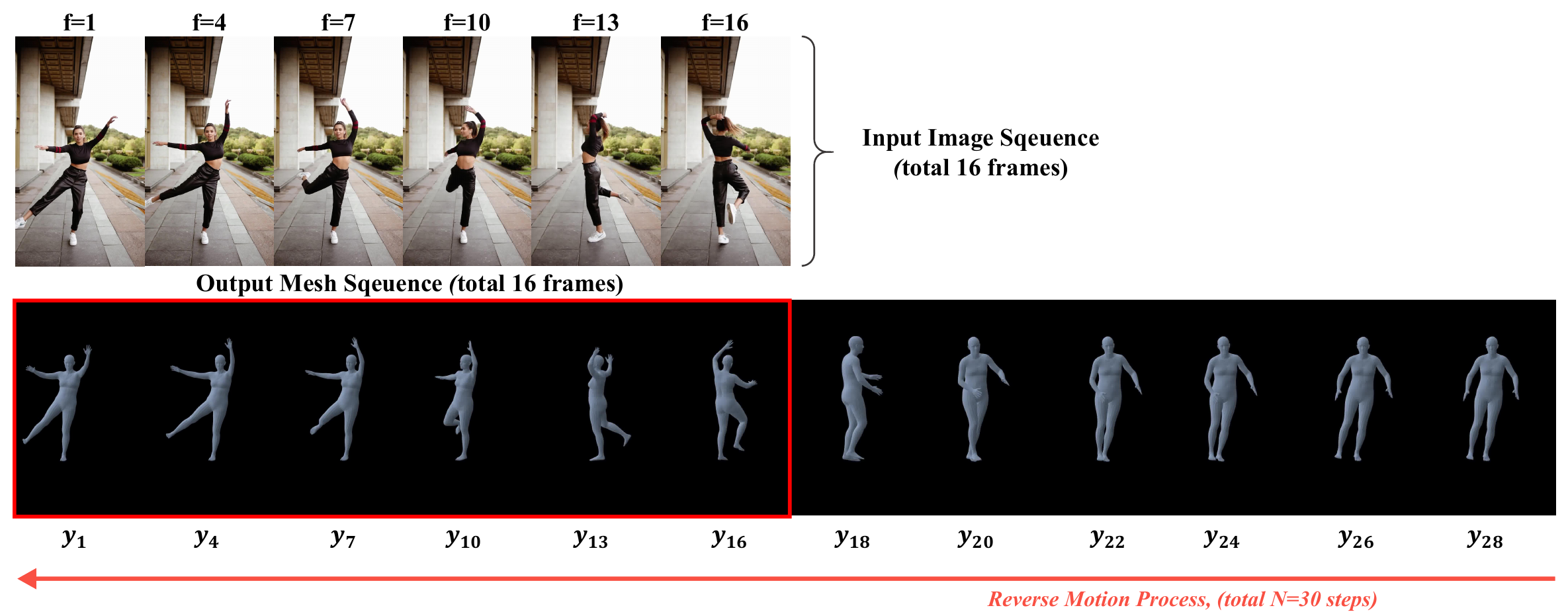}
  \vspace{-5pt}
  \caption{Visualization of decoding steps during the reverse motion process.}
  \label{fig:supp_full_frame}
  \vspace{-5pt}
\end{figure*}

In our experimental setup, we utilize 16 input frames, and the total number of steps is set to 30. In the reverse motion process, DiffMesh outputs [$y_1, y_2 \cdots, y_{30}$] over 30 steps. For the output mesh sequence of 16 frames, we use [$y_1, y_2 \cdots, y_{16}$]. Additionally, we generate the mesh from [$y_{16}, y_{17} \cdots, y_{30}$], as visually depicted in Fig \ref{fig:supp_full_frame}. This visualization illustrates the trend of the generated human mesh gradually decoding toward the desired human mesh of each input frame.

\begin{figure*}[htp]
\vspace{-10pt}
  \centering
  \includegraphics[width=1\linewidth]{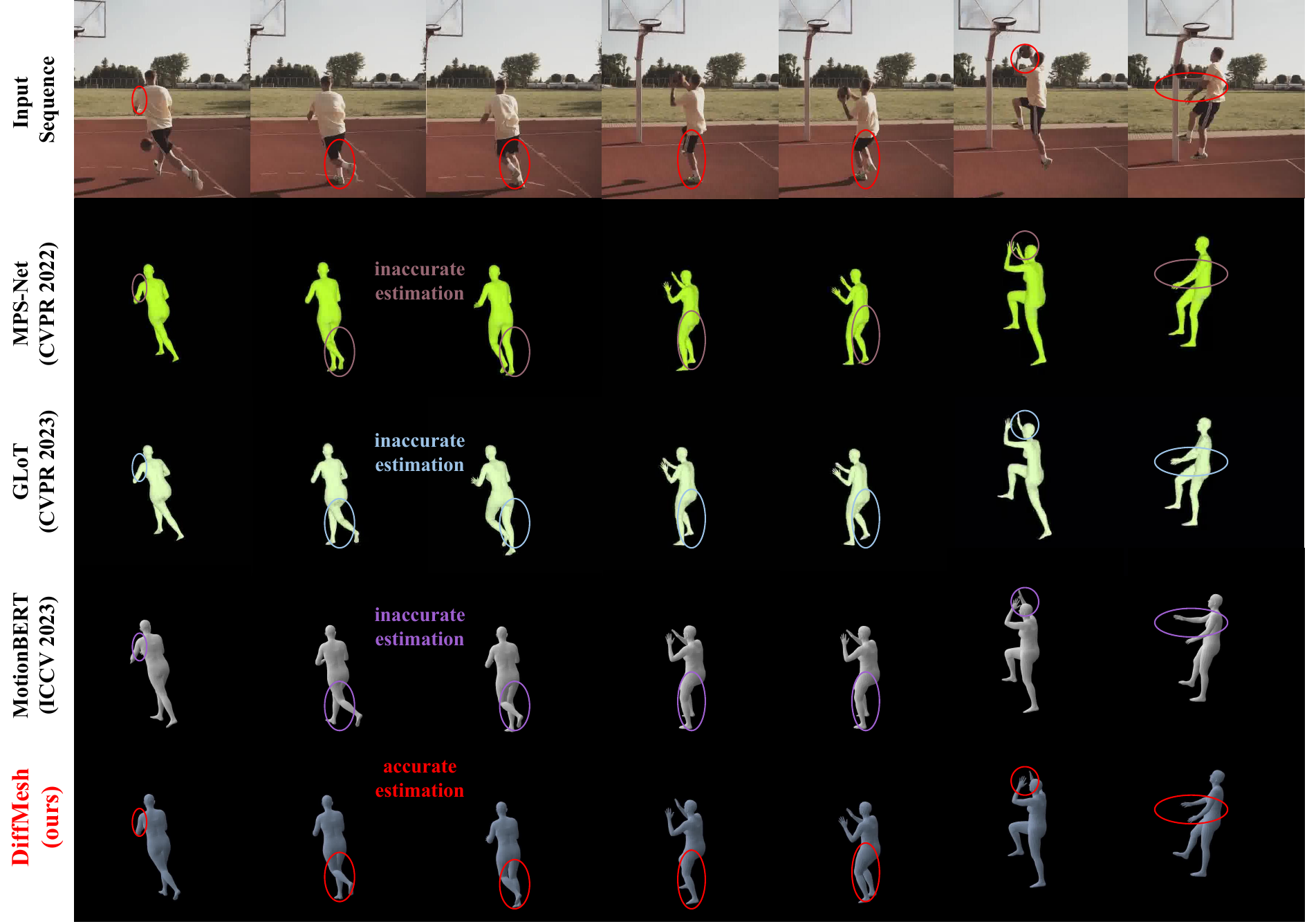}
  \vspace{-5pt}
  \caption{Other qualitative results of our DiffMesh on in-the-wild videos. \textcolor{blue}{Please refer to our \textbf{video demo} for the more reconstructed mesh sequences results.}}
  \label{fig:supp_vis2}
  \vspace{-10pt}
\end{figure*}

Furthermore, we show the qualitative results of DiffMesh on \textbf{in-the-wild videos} in Fig.~\ref{fig:supp_vis2}. We observe that DiffMesh demonstrates remarkable performance in reconstructing more reliable human mesh sequences with temporal consistency compared to previous methods. \textcolor{blue}{Please refer to our \textbf{video demo} for the more reconstructed mesh sequence results.}

\section{Broader impact and limitation}
\label{Broader}
DiffMesh establishes an innovative connection between diffusion models and human motion, facilitating the generation of accurate and temporal smoothness output mesh sequences by integrating human motion into both the forward and reverse processes of the diffusion model. By enabling direct 3D human mesh reconstruction from 2D video sequences, DiffMesh eliminates the dependency on additional motion sensors and equipment, thereby streamlining the process and reducing costs.

However, despite its advancements, DiffMesh is not without limitations. Similar to previous methods, DiffMesh may face challenges in scenarios with substantial occlusions, resulting in the production of unrealistic mesh outputs. To address this issue, further exploration into spatial-temporal interactions within the human body is warranted, serving as a focal point for our future research. Additionally, DiffMesh may encounter difficulties in rare and complex pose scenarios due to the constraints of limited training data, highlighting the necessity for ongoing development and refinement efforts.

\clearpage
\newpage

{\small
\bibliographystyle{ieee_fullname}
\bibliography{egbib}
}

\end{document}